\documentclass[]{hdsr}
\graphicspath{{figs/}}
\usepackage{subcaption}
\usepackage[]{placeins}
\usepackage{yhmath}

\newcommand{\R}{\mathbb R}
\newcommand{\cal}[1]{\mathcal{#1}}
\newcommand{\ev}[1]{\mathbb E \left [#1\right ]}

\begin{document}

\newgeometry{bottom=1.5in}

\volumeheader{0}{0}{00.000}

\begin{center}

  \title{Revisiting Broken Windows Theory}
  \maketitle

  \thispagestyle{empty}
  
  \vspace*{.2in}

  \begin{tabular}{cc}
    Ziyao Cui\upstairs{\affilone}$^*$, Erick Jiang\upstairs{\affilone}$^*$, Nicholas Sortisio\upstairs{\affilone}$^*$, Haiyan Wang\upstairs{\affilone}$^*$, \\
    Eric Chen\upstairs{\affilone}, Cynthia Rudin\upstairs{\affilone}
   \\[0.25ex]
   {\small \upstairs{\affilone} Department of Computer Science, Duke University, Durham, NC} \\
  \end{tabular}
  
  \emails{
    \upstairs{*}richard.cui@duke.edu, erick.jiang@duke.edu, nicholas.sortisio@duke.edu, haiyan.wang916@duke.edu, eric.y.chen@duke.edu, cynthia@cs.duke.edu
    }
  \vspace*{0.4in}

\def\thefootnote{*}\footnotetext{Equal contribution.}

\begin{abstract}
We revisit the longstanding question of how physical structures in urban landscapes influence crime. Leveraging machine learning-based matching techniques to control for demographic composition, we estimate the effects of several types of urban structures on the incidence of violent crime in New York City and Chicago. We additionally contribute to a growing body of literature documenting the relationship between perception of crime and actual crime rates by separately analyzing how the physical urban landscape shapes subjective feelings of safety. Our results are twofold. First, in consensus with prior work, we demonstrate a ``broken windows'' effect in which abandoned buildings, a sign of social disorder, are associated with both greater incidence of crime and a heightened perception of danger. This is also true of types of urban structures that draw foot traffic such as public transportation infrastructure. Second, these effects are not uniform within or across cities. The criminogenic effects of the same structure types across two cities differ in magnitude, degree of spatial localization, and heterogeneity across subgroups, while within the same city, the effects of different structure types are confounded by different demographic variables. Taken together, these results emphasize that one-size-fits-all approaches to crime reduction are untenable and policy interventions must be specifically tailored to their targets.
\end{abstract}
\end{center}

\vspace*{0.15in}
\hspace{10pt}
  \small	
  \textbf{\textit{Keywords: }} {Causal Inference, Broken Windows Theory, Perceived Danger, Matching}
\copyrightnotice

\section*{Media Summary}

Researchers and policymakers have long been interested in identifying and characterizing the relationship between the built urban environment and crime. One such theory is the Broken Windows Theory (BWT), which suggests that signs of social disorder such as abandoned buildings and broken windows encourage crime in surrounding areas. Focusing on crime in Chicago and New York City, this study reinforces and expands upon BWT in three ways. First, we revisit the original hypothesis of BWT by analyzing the criminogenic effects of abandoned buildings. Second, we show that the effects described by BWT extend to a broader class of urban structure types: those that draw foot traffic (e.g. libraries, schools, and public transport). Third, we ask whether effects on actual crime are reflected in perceptions of crime and safety. Our results support BWT, with abandoned buildings being associated with higher incidences of crime in surrounding areas. We demonstrate a similar effect for structure types with heavy foot traffic, and we show that this effect is associated with a corresponding increase in perceived danger throughout surrounding communities. However, these effects are highly specific. When comparing across cities, the effect of the same structure type is different in magnitude and degree of localization while also experiencing heterogeneity with respect to different subgroups. Within cities, the impacts of different structure types are affected by different demographic characteristics. Taken together, our results indicate that policy interventions focused on the design of urban public spaces are a promising approach to crime reduction, but these interventions must be carefully tailored to the unique spatial, structural, and social dynamics of each locality.

\section{Introduction}
\label{sec1}

Broken Windows Theory (BWT) hypothesizes that signs of social disorder such as abandoned properties, rampant graffiti, or broken windows incentivize crime in surrounding areas \citep{wilson1982broken}.  Empirical studies have yielded mixed results on the association between social disorder and crime \citep{welsh2015meta}, but BWT has nevertheless motivated a number of new policing strategies in various American cities aimed at reducing the prevalence of these indicators. Unsurprisingly, these strategies, many of which aggressively pursue misdemeanors like vandalism in an attempt to maintain ``order'' \citep{harcourt2015BWtalk}, have proven controversial, with disagreement arising over their effectiveness \citep{braga2008spots, manhattan2001police, harcourt2006experiment}. 

BWT is part of a broader prevailing interest in the relationship between urban planning and crime \citep{newman1972defense, skogan1990}. Specific structure types within urban environments such as green space \citep{macdonald2021reducing}, late-night bars \citep{burgason2017close}, and public parks \citep{groff2012role} have all been found to exert substantial effects on surrounding crime rates. However, these effects are highly variable both within and between cities \citep{connealy2020can}, and many urban structure types have yet to be studied. Additionally, these studies ignore an important dimension of criminogenesis: the psychological factor. Wilson and Kelling note the significance of public perception in the emergence of crime, writing

\begin{quote}
    ``...many residents will think that crime, especially violent crime, is on the rise, and they will modify their behavior accordingly. They will use the streets less often, and when on the streets will stay apart from their fellows, moving with averted eyes, silent lips, and hurried steps... In response to fear people avoid one another, weakening controls'' \citep{wilson1982broken}.
\end{quote}

Given these gaps in knowledge and their potential importance to criminal justice policy, we aim to further explore the relationship between urban environments and crime. This study makes three methodological improvements on previous studies. (1) We make use of larger and more comprehensive datasets: government-sponsored spatiotemporal datasets concerning urban planning, crime, and population demographics paint a clearer picture of modern cities than previously available. (2) We apply a modern matching method for treatment effect estimation in an observational setting. Crucially, these methods exceed the performance of state-of-the-art black box methods while being \emph{interpretable}, yielding high-quality and trustworthy matches while providing previously unavailable insight into confounding effects. (3) We ask how the urban environment contributes to perceptions of safety, a previously unaddressed dimension of BWT that cannot be captured through analysis of crime rates alone.

Our work presents the following results:\footnote{The code base for our investigation is available on \href{https://github.com/richardcui18/revisiting-bwt}{GitHub} for future reference and development.}
\begin{enumerate}
    \item Abandoned buildings are associated with higher crime and perceived danger, corroborating BWT. Additionally, while the former effect is observable in census tract-level crime rate data, much of the effect is highly localized: crime density is observably elevated within meters of the building in question. A positive association with crime and perceived danger is also observed for structure types that draw higher levels of foot traffic.
    \item The effects of these structure types on crime are not uniform within or between cities. The effects of different structure types within the same city are confounded by different demographic variables, while the effects of the same structure type across cities are not uniform in magnitude, degree of localization, or heterogeneity across subgroups.
\end{enumerate}

Our paper proceeds as follows: Section \ref{sec:r_works} reviews previous work on the relationship between crime and the urban environment. Section \ref{sec:methodology} presents our methodology. Section \ref{sec:results} presents the results of our analysis on data from New York and Chicago. Section \ref{sec:conclusion} discusses our conclusions and analysis.

\section{Related Work}
\label{sec:r_works}

\subsection{Urban Landscapes and Crime}
\label{subsec:r_works_urban}

In her seminal work \textit{Death and Life of Great American Cities}, Jane Jacobs theorized that public order could be maintained through urban design that encouraged more ``eyes on the street;'' thus, more vital, heavily trafficked areas create a self-reinforcing cycle of crime prevention \citep{jacobs1961death}. Oscar Newman's defensible space theory also centered on the role of urban design as a crime prevention tool, but instead posited that densely trafficked areas contributed to a sense of anonymity that would in turn lead to increased levels of crime \citep{newman1972defense}. 

Following these contradictory theoretical advances, BWT emerged in the 1980s, and it has remained a prominent topic of criminology research since its inception \citep{wilson1982broken}. It gained traction for its intuitive link between disorder and crime, offering a clear framework for both theory and practice. A number of explanations drawing on fields ranging from economics to social psychology have suggested causes for BWT: perceived breakdowns in social norms and systems of control, disparities in socioeconomic status and geographic mobility, and changes in policing strategies are all cited as potential contributors \citep{wilson1982broken}. Early work on BWT primarily applied statistical analysis to assess its theoretical predictions, with some results supporting its conclusions \citep{manhattan2001police, corman2005carrots}, while others contradicted them \citep{harcourt2006experiment, harcourt2001illusion}.

A large body of more recent work has focused on the Philadelphia LandCare project, an ongoing initiative to green abandoned lots and encourage urban forestry. Existing work has exploited quasi-experimental variation in the selection of lots for greening under the program in order to explore how vacant lot remediation could reduce nearby crime. Most related to our work are \citet{branas2018citywide}, \citet{macdonald2021reducing}, and \citet{cui2022matching}. 
\citet{branas2018citywide} found that greening vacant lots led to reduced crime in high-poverty areas and increased feelings of safety among nearby residents. 
\citet{macdonald2021reducing} used an entropy distance weighting framework to match greened and ungreened lots within census tracts and found that the moderating effects of lot greening are highly dependent on other nearby land uses. \citet{cui2022matching} used propensity score matching to control for differences in the surrounding environments of greened and ungreened lots across tracts and found a 2-3\% reduction per year of crime around greened lots. In a related study, \citet{south2018effect} used survey data from a cluster-randomized lot-greening program, and found that residents near remediated lots reported much better mental health than the non-intervention baseline. However, these types of interventions are extremely costly and, as a result, are unlikely to be carried out across a representative set of localities. Similar initiatives are underway in Detroit and Flint, Michigan, New Orleans, Louisiana \citep{kondo2018blight}, and Youngstown, Ohio \citep{kondo2016effects, heinze2018busy}, and although each program offers further opportunities to study the effects of greening policies in more locations, they are concentrated in cities already suffering from wide-scale urban decay. This concentration provides valuable evidence for contexts where abandonment is most severe, but it also raises questions about how far the findings generalize to healthier or more rapidly growing urban environments.

Previous machine learning approaches in this area typically treat the relationship between the urban landscape and crime as a forecasting problem \citep{wheeler2021, zhang2020, kim2018}. As a result, many of these works use crime data as predictors of future crimes \citep{flaxman2019scalable, rotaru2022event}, or rely on non-interpretable deep learning techniques \citep{kang2017prediction, stec2018forecasting}, which can be especially problematic since they may introduce racial or socioeconomic biases that cannot be assessed due to the black-box nature of these models. Within this literature, our study is most closely related to \citet{nadai2020socio}, who construct a Bayesian model of neighborhood crime based on the local urban and socioeconomic environment. They use data on land use and mobility patterns for a number of cities across several countries and find that certain signals of urban vibrancy (e.g., stores, restaurants, pedestrian traffic, etc.) are positively related with crime, cutting against the predictions of Jacobs' theory and BWT. However, much of their model is not replicable, as it is built on non-public cellphone records, while the approach we present here is based entirely on publicly available data.

\subsection{Perceived Danger}
\label{subsec:r_works_perceived}

Recent survey data have revealed a sizable contrast between Americans' perceptions of the level of crime and the levels of actual crime borne out by crime data \citep{gallup2023more}. Existing analysis of this trend have tended to focus on partisan affiliation \citep{gallup2022record} and media consumption \citep{pewlocalcrime}. In particular, \citet{gallup2022record} and \citet{gallup2023more} report that self-identifying Republicans report perceived rates of violent crime substantially higher than those reported by Democrats, although both groups overestimate the true crime rate. More recently, \citet{pewlocalcrime} found that consumption of local news coverage of crime is associated with increased perceived victimization risk and the belief that crime should be a top policy priority. However, no prior study that we know of has tried to directly link urban structure types with self-reported \emph{perceived} danger. That is, no one has previously tried to evaluate BWT and/or Jane Jacob's theory as explanations for perceived danger.

\subsection{Our contributions}
\label{subsec:r_works_contributions}

 Our work differs from the existing literature in several key respects. In particular, we estimate how aspects of the built environment (structures such as public transportation infrastructure and abandoned buildings) contribute to perceived danger separately from their influence on actual levels of crime, and thus can directly identify factors that influence perceived danger more strongly than actual crime or vice-versa. Both are of independent interest if policymakers are jointly interested in actual and perceived levels of safety. Our work is also the first to apply the recently developed almost-matching-exactly framework \citep{WangEtAlFLAME2021, parikh2022malts} to this area of research, and we do so on a much larger set of demographic variables than previous studies on BWT. Moreover, we do so in a flexible framework that allows for meaningful inter-city comparison, while most previous work focuses on a single locale \citep{macdonald2018schools, bernasco2008robberies, braga2008spots} or only considers cities of interest separately without inter-city comparison \citep{barnum2017kaleidoscope, skogan1977changing}. We gain substantial generalizability from incorporating many relevant and generalizable public data sources; we use these data and interpretable causal inference techniques to produce meaningful conditional average treatment effects. Both of these are especially important in light of well-documented treatment effect heterogeneity within and across urban settings. 

\section{Methodology}
\label{sec:methodology}
We analyzed the effects of several ubiquitous urban structure types on the incidence of violent crime and perceptions of safety across Chicago and New York City between 2008-2022. Structures were selected to re-evaluate BWT and to examine how crime is influenced by high levels of foot traffic. Using US census tracts as the unit of observation, as they generally have comparable population sizes, we applied recently developed supervised and unsupervised statistical matching techniques to define treatment and control groups for robust observational causal inference. Specifically, tracts were matched on their demographic composition, which prior research finds to be highly correlated with crime \citep{entorf2000socioeconomic, stucky2016intra}. Within matched groups, we aimed to explore crime patterns at varying spatial resolutions, capturing broad trends at the census tract level as well as highly localized trends in close proximity to the structures in question. To do so, we considered two measures of crime: crime rate per capita (aggregated at the tract level) and crime density per square mile (aggregated over concentric regions centered on structures of interest).

Section \ref{subsec:data} summarizes our data collection and processing methods; Section \ref{subsec:treatments} explains the treatments we considered; Section \ref{subsec:matching} introduces our matching procedure; Section \ref{subsec:ate} defines average and conditional average treatment effects; and Section \ref{subsec:rate_vs_density} defines our treatment effect measures: crime rate and density, respectively.

\subsection{Data}
\label{subsec:data}

Demographic data were provided by the U.S. Census Bureau's American Community Survey (ACS). We considered demographic composition at the census tract level during three consecutive five-year periods: 2008-2012 (S1), 2013-2017 (S2), and 2018-2022 (S3) \citep{ACS5year}.  We elected to use multiyear estimates due to their larger sample sizes which enhance statistical reliability, particularly for small geographic regions with low populations. The 31 demographic variables are listed in Appendix \ref{appendix:demographic variables} and encompass population size, age, race, sex, educational attainment, socioeconomic status, and family structure. Unpopulated census tracts were excluded. Shapefiles describing the geography of these tracts were also publicly available through the Census Bureau \citep{censusGeo}.

Urban planning data were available through the New York City and Chicago open data portals. We considered seven structure types: abandoned buildings, bus stops, subway stations (rail stations in Chicago), libraries, public schools, restaurants, and grocery stores. Every dataset provided exact geographic coordinates for each structure. Some datasets also contained information about when a structure was built or opened. These dates were considered in our analysis when available, otherwise we assumed every structure was present since the beginning of our analysis period in 2008. 

The New York City (NYPD) and Chicago (CPD) police departments both release up-to-date crime data including the exact geographic coordinates where the crime allegedly occurred. Chicago crime data, compiled by the CPD's Citizen Law Enforcement Analysis and Reporting system, encompasses all reported crimes in the city since 2001 \citep{chicago2024crime}. New York City crime data, compiled by the NYPD, contains all reported felonies, misdemeanors, and violations since 2006 \citep{nyc2024crime}. We analyzed reported crimes as opposed to arrests to mitigate the impact of external factors such as policing strategies. From these raw datasets, we selected for violent crimes -- battery, rape, and homicide.

Perceived danger is less well documented, and we were only able to access reliable data from Chicago via the Healthy Chicago Survey (HCS) \citep{healthychicago} which provided polling data on perceptions of safety 
in 2021 and 2022 aggregated over Chicago's 77 Community Areas. We specifically considered their perceived neighborhood violence rate data,  which reports the ``percent of adults who reported that violence occurs in their neighborhood `every day' or `at least every week.'''

\subsection{Defining Treatments}
\label{subsec:treatments}

The matching methods to be introduced in Section \ref{subsec:matching} require binarized treatments. This binarization is not always natural when considering urban structures. Some structure types such as libraries are naturally binarized or ``sparse'' -- they are distributed such that every census tract contains at most one such structure while many tracts contain none. However, ``dense'' structure types such as bus stops are distributed such that every tract contains at least one if not multiple instances of the structure. In this case, we introduced a ``treatment threshold'' on the number of structures in each tract. Above this threshold, a tract was considered treated; at or below it, the tract was considered untreated.

We considered two factors when setting treatment thresholds: balancing experimental group size and minimizing the potential confounding effect of control tracts that still contain the structure type in question. To account for both, we set the treatment threshold of all dense structures to be the median across tracts, guaranteeing approximately equally-sized treatment and control groups. However, we dropped tracts in the fourth and fifth deciles to reduce the potential confounding effect of ``borderline'' tracts near the threshold.

\subsection{Matching}
\label{subsec:matching}

We applied the Matching After Learning to Stretch (MALTS) algorithm to generate matched groups for treatment effect estimation. MALTS learns a weighted Euclidean distance metric over the covariate space \citep{parikh2022malts}, ensuring that relevant covariates (i.e., those that are highly correlated to the treatment outcome) contribute more to the distance between points while irrelevant covariates barely contribute at all. Units are matched to their nearest neighbors according to this learned metric. 

\begin{figure}[ht]
    \centering
    \includegraphics[width = 0.75\linewidth]{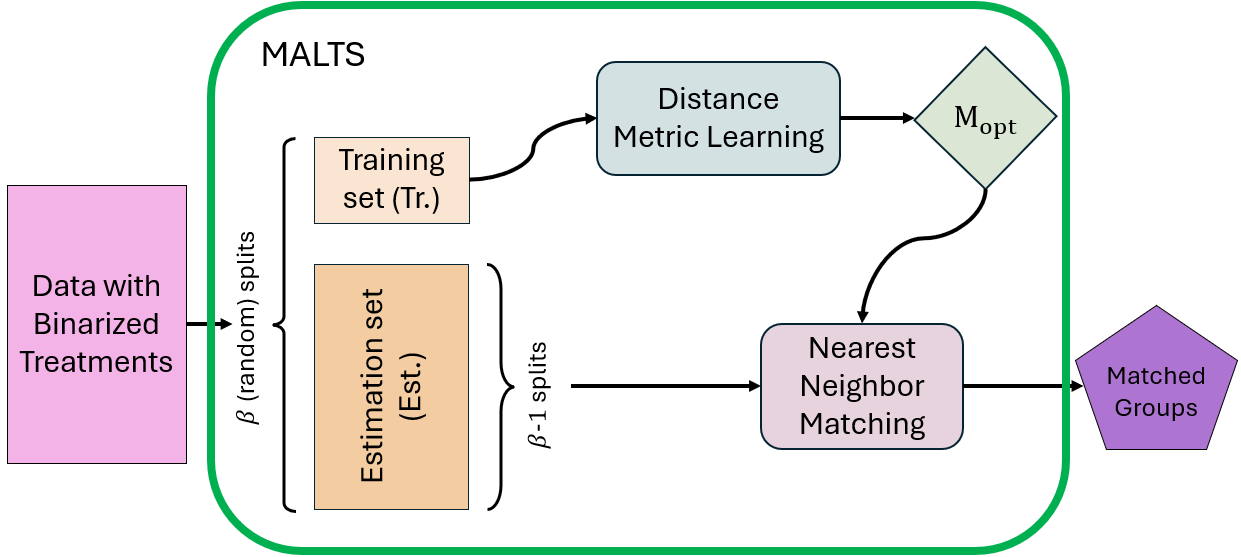}
    \caption{MALTS Framework \citep{parikh2022malts}}
    \label{fig:malts_framework}
\end{figure}

We used MALTS to match census tracts by their demographic composition (see Section \ref{subsec:data} and Appendix \ref{appendix:demographic variables}) to eliminate any confounding effect demographics may have on crime. Rather than splitting our data into a single training and estimation fold, we generated matched groups with repeated cross-validation where two units were considered matched if they were matched in multiple folds. This reduced the likelihood of a biased treatment effect estimate due to a flawed distance metric and increased the amount of data available for estimation. A diagram of our MALTS framework, based on the figure of \citet{parikh2022malts}, is presented in Figure \ref{fig:malts_framework}. 

In comparison to other matching methods, MALTS offers two significant advantages that we exploited in this study. (1) It matches ``almost exactly,'' meaning matched units are similar in their covariates. This allowed us to estimate \emph{conditional} average treatment effects that provide insight into treatment effect heterogeneity. (2) It is interpretable, returning a simple distance metric that enabled us to better understand how units were being matched and to evaluate the quality of each matched group. Specifically, we followed the procedure proposed by \citet{parikh2022malts} where, for each matched group, we computed its diameter, or the distance from the unit being matched to the furthest unit in its group according to the learned metric. Matched groups with large diameters were dropped, further ensuring match quality and estimation accuracy.

\subsection{Treatment Effect Estimation}
\label{subsec:ate}

The average treatment effect (ATE) is a statistical estimation of the effect of a treatment relative to a control group. ATE estimates average out any heterogeneity in treatment effect and reflect only the mean difference between treated and control units. On the other hand, the conditional average treatment effect (CATE) estimates model treatment effects for a \textit{specific subgroup}. In the context of our study, this will often be a group of census tracts sharing similar demographic compositions. In this section, we define our methods for ATE and CATE estimation more rigorously.

Let $\cal X \subset \R^p$ denote a $p$-dimensional covariate space, $\cal Y \subset \R$ denote the outcome of interest, and $\cal T$ denote a binary treatment indicator. Let $\cal Z = \cal X \times \cal Y \times \cal T$ represent the data space and let $\cal S_n \subset \cal Z$ be an observed set of $n$ samples.  For a single observation $z_i = (x_i, y_i, t_i) \in \cal S_n$, let $y_i^{(T)}$ be its potential outcome with treatment and  $y_i^{(C)}$ its potential outcome without treatment. The CATE on $z_i$ is defined as the expected difference in the unit's outcome under the treatment and control:
\begin{align*}
    \text{CATE}(z_i) = \ev{y_i^{(T)} - y_i^{(C)} \mid x = x_i} = \ev{y_i^{(T)} \mid x = x_i} - \ev{y_i^{(C)} \mid x = x_i}.
\end{align*}
The ATE follows as the expected CATE over the full data distribution: 
\begin{align*}
    \text{ATE} = \mathbb E_{z \in \cal Z} [\text{CATE}(z)].
\end{align*}
We cannot observe true values of $\ev{y_i^{(T)} \mid x = x_i}$ or $\ev{y_i^{(C)} \mid x = x_i}$, so we instead estimate them with the matched group of unit $i$. Let $MG^{(T)}_i$ be the treated units matched with $i$ and $MG^{(C)}_i$ the untreated units matched with $i$. For some estimator $\phi$, the estimated CATE at $x_i$ is given by
\begin{align*}
    \widehat{\text{CATE}}(x_i) = \phi \left( MG_i^{(T)} \right)  - \phi \left( MG_i^{(C)} \right).
\end{align*}
We opted for the difference-in-means approach with the estimator given by $\phi(MG_i) = \frac{1}{|MG_i|} \sum_{z_j \in MG_i} y_j$, where each matched unit is denoted $z_j=(x_j,y_j,t_j)$ and $y_j$ is its observed outcome. As before, the estimated ATE is given by
\begin{align*}
    \widehat{\text{ATE}} = \mathbb E_{z \in \cal S_n} [\widehat{\text{CATE}}(z)].
\end{align*}
Lastly, as suggested by \citet{parikh2022malts}, we estimated the variance of each CATE estimate with gradient-boosted quantile regression. 

\subsection{Crime Metrics: Tract-Level Rate and Localized Density Analysis}
\label{subsec:rate_vs_density}

A key contribution of our study is the analysis of crime patterns at multiple spatial resolutions. We considered treatment effects at both census tract level and in smaller, localized regions around structures. At the census tract level, we estimated treatment effects with respect to tract-level crime rate. Over the smaller regions, we estimated treatment effects with respect to crime density. 

Our tract-level analysis follows directly from the matching and treatment effect estimation described above. We used MALTS to match census tracts by their demographic composition with crime rate as the outcome of interest. With these matched groups, we estimated the CATE of each structure on every tract and aggregated these estimates into the overall ATE of the structure type.

We were additionally interested in whether the structure types in our study exert highly localized effects on crime that are drowned out once aggregated with the rest of the tract. The geographic specificity of our crime data allowed us to answer this question, and we estimated the treatment effect of structure types on crime in increasingly large regions around them. However, this spatial resolution necessitated us to analyze effects on crime \emph{density}, defined as the number of crimes in a given area over a period of time. This change is due to the lack of more geographically specific population data, which precluded calculating crime rates in areas smaller than tracts.
We again generated matched groups with MALTS using the tract-level crime rate as the outcome. For treated units in each matched group, we calculated the mean crime density in concentric circular regions centered on each structure. The radii of the regions are set by the user, and we elected to analyze 16 radii beginning at 25m with 25m increments between them. For untreated units, we sampled a preset number of points from each tract and calculated the mean crime density in the same concentric regions centered at these points. This sampling procedure consists of randomly selecting a user-specified number of points within a user-specified radius around each tract’s centroid. In our study, we sampled 20 random points within a 750 m radius of each control tract’s centroid.

\section{Results}
\label{sec:results}

In Section \ref{subsec:bwtres}, we ask whether BWT generalizes across different cities and over different time periods by estimating the treatment effects of abandoned buildings on surrounding crime. Section \ref{subsec:ftres} does the same for urban structure types that draw high levels of foot traffic. Section \ref{subsec:perceivedres} demonstrates how these structure types impact perceptions of safety and addresses how these perceptions compare to actual crime. Section \ref{subsec:heterogeneity} analyzes these results more granularly, using the relationship between de-aggregated CATE estimates and selected covariates to better understand treatment effect heterogeneity over different demographic subgroups.

\subsection{BWT Holds for Abandoned Buildings}
\label{subsec:bwtres}

Our analysis focuses on abandoned buildings reported in 311 calls to the city government. This is potentially the most reliable sign of social disorder for which data is up-to-date and collected across multiple cities. 

\begin{figure}[h]
    \centering
    \begin{subfigure}[b]{0.48\textwidth}
        \centering
        \includegraphics[width=\textwidth]{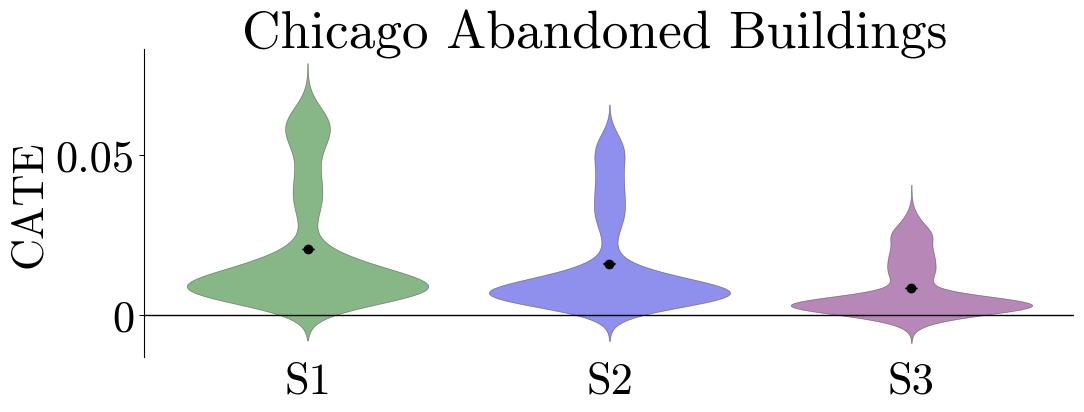}
        \label{fig:mtl_chicago_abandoned_buildings}
    \end{subfigure}
    \hfill 
    \begin{subfigure}[b]{0.48\textwidth}
        \centering
        \includegraphics[width=\textwidth]{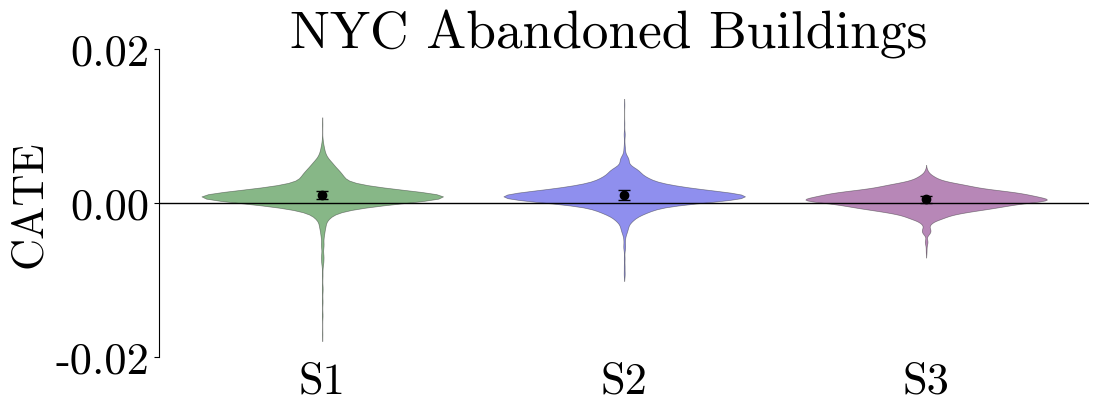}
        \label{fig:mtl_nyc_abandoned_buildings}
    \end{subfigure}
    \caption{\textit{Tract-level analysis shows the effect of abandoned buildings on crime is pronounced over broader geographic regions}. We see positive CATE distributions for the effect of abandoned buildings over the three time frames, with ATE estimates plotted as dots in the figure being above 0 indicating the presence of abandoned buildings in tracts being related to higher crime rates. Also, this effect is more prominent in Chicago and appears to weaken in S3.}
    \label{fig:mtl_ab} 
\end{figure}

\begin{figure}[h]
    \centering
    \begin{subfigure}[b]{0.48\textwidth}
        \centering
        \includegraphics[width=\textwidth]{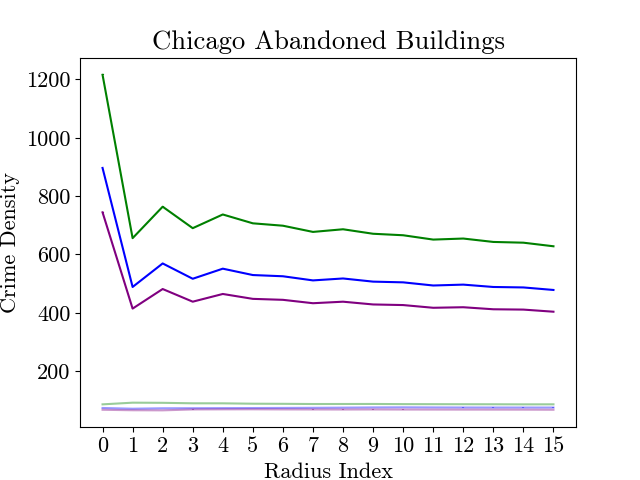}
        \label{fig:mt_chicago_abandoned_buildings}
    \end{subfigure}
    \begin{subfigure}[b]{0.48\textwidth}
        \centering
        \includegraphics[width=\textwidth]{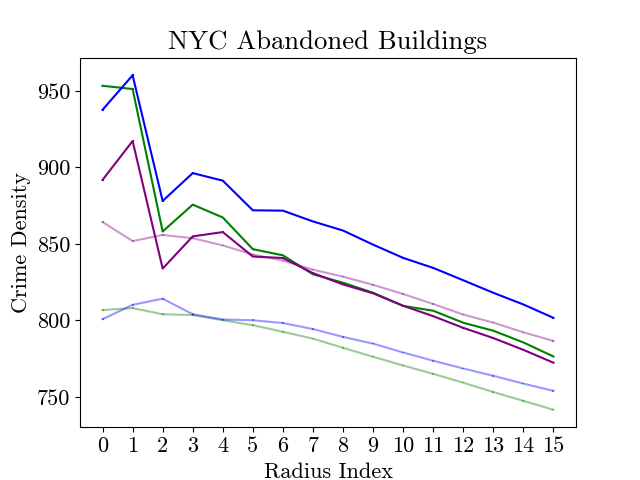}
        \label{fig:mt_nyc_abandoned_buildings}
    \end{subfigure}
    \begin{subfigure}[b]{0.6\textwidth}
        \centering
        \fbox{\includegraphics[width=\textwidth]{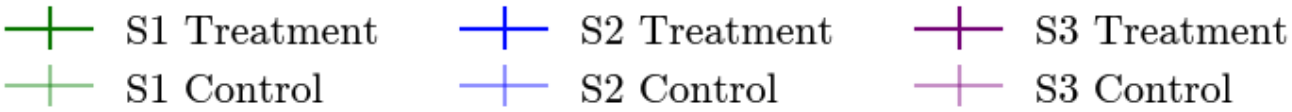}}
    \end{subfigure}
    \vspace{0.1in}
    \caption{ \textit{Density analysis in Chicago and New York shows the effect of abandoned buildings on crime density is highly localized}. We see larger differences between treatment and control outcomes at small radii that decrease rapidly with distance. However, these effects differ across both cities and analysis periods: in Chicago, the effect decreases in later treatment periods, but crime density in treated tracts is much higher than in the control. The same is not true of New York. It should be noted that the two plots are presented on different scales; thus, changes in the control sequence in New York are still relatively small, as expected, and comparable to those in Chicago.}
    \label{fig:mt_ab} 
\end{figure}

Results shown in Figures \ref{fig:mtl_ab} and \ref{fig:mt_ab} agree with classical BWT: abandoned buildings have a substantially positive effect on crime rate. As seen in Figure \ref{fig:mt_ab}, much of this difference is localized, with crime density spiking in close proximity to abandoned buildings and decreasing with distance. These effects are not consistent between cities. In Chicago, all three treatment series decrease rapidly with distance toward a flat asymptote that is consistently above the control series, indicating the structure type consistently has a tract-level effect on crime. In contrast, only the treatment series for periods S1 and S2 in New York remain consistently above the control; the treatment series for S3 is not much different from the control series, dipping below the control. This is reflected in the tract-level ATE estimates in Figure \ref{fig:mtl_ab}, with abandoned buildings in Chicago exerting stronger treatment effects compared to those in New York. These effects are not even consistent across analysis periods within the same city. In Chicago, the effect of abandoned buildings dissipates during S3 at both localized and regional levels. While our results agree with classical BWT, we have shown a nuanced local and global relationship between abandoned buildings and crime that can vary with city or time period.

\subsection{Foot Traffic Induces Higher Crime Rates}
\label{subsec:ftres}

We analyze the treatment effects of urban structure types that draw substantial foot traffic and ask whether certain kinds of foot traffic are more strongly correlated with crime. 

\begin{figure}[h]
    \centering
    \begin{subfigure}[b]{0.48\textwidth}
        \centering
        \includegraphics[width=\textwidth]{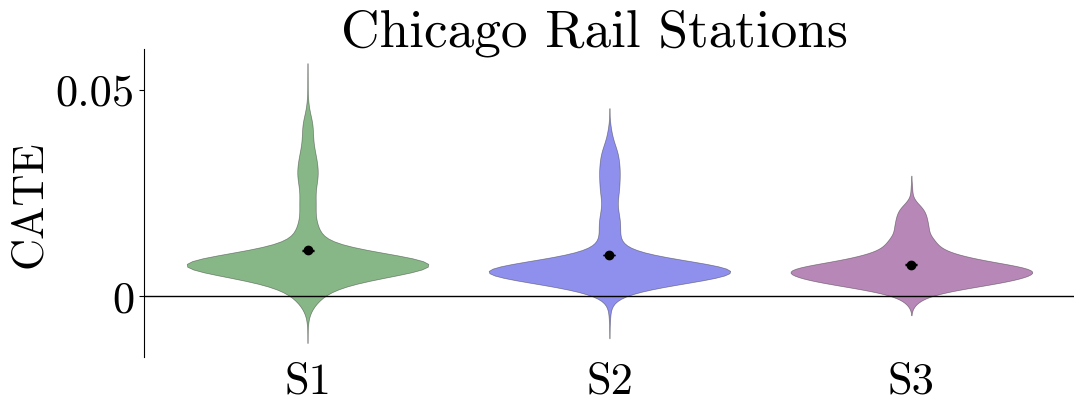}
        \label{fig:mtl_chicago_rail_stations}
    \end{subfigure}
    \hfill
    \begin{subfigure}[b]{0.48\textwidth}
        \centering
        \includegraphics[width=\textwidth]{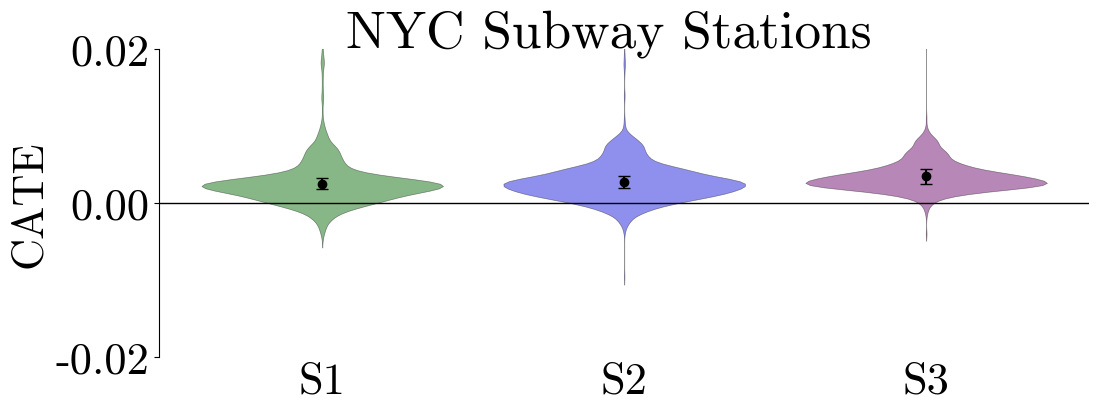}
        \label{fig:mtl_chicago_bus_stops}
    \end{subfigure}
    \begin{subfigure}[b]{0.48\textwidth}
        \centering
        \includegraphics[width=\textwidth]{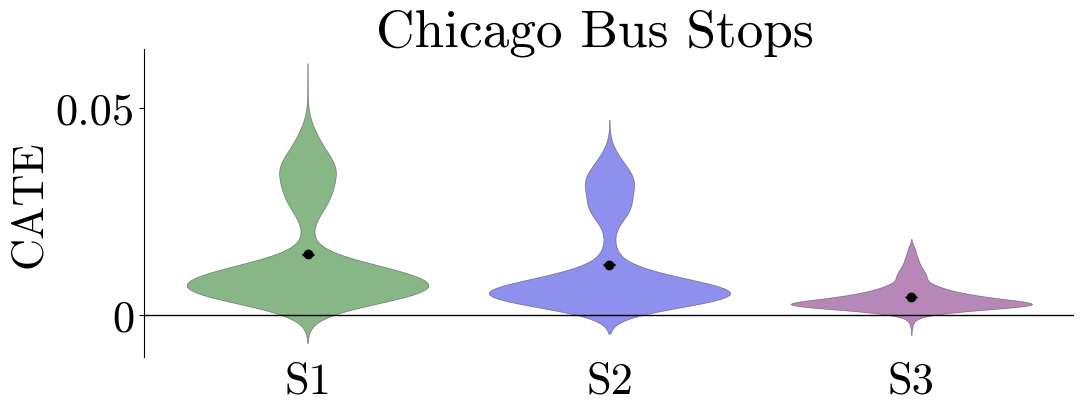}
        \label{fig:mtl_nyc_bus_stops}
    \end{subfigure}
    \hfill
    \begin{subfigure}[b]{0.48\textwidth}
        \centering
        \includegraphics[width=\textwidth]{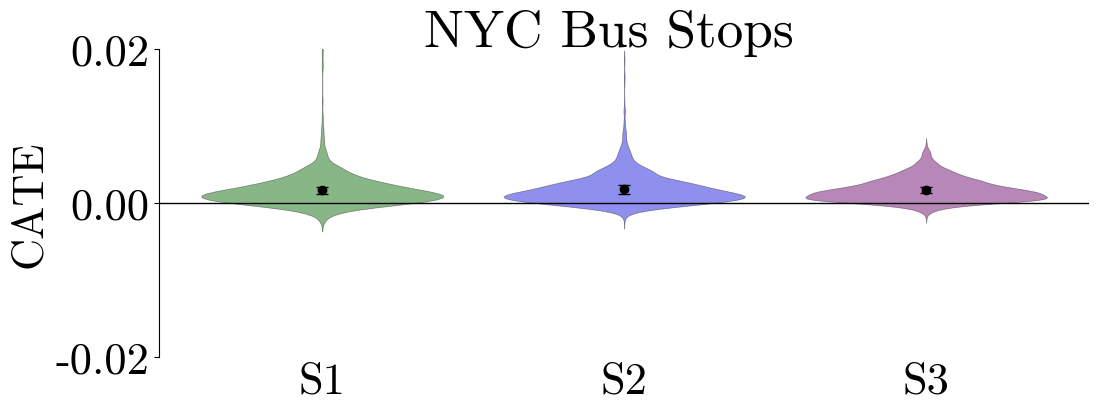}
        \label{fig:mtl_nyc_subway_stations}
    \end{subfigure}
    \caption{
    \textit{There is more crime in tracts with public transportation infrastructure}: these structures all have positive effects on crime, with the CATE distributions being mainly positive, though these effects are more pronounced in Chicago. The effect of these structures in Chicago also decreases in S3.}
    \label{fig:mtl_pubtransport}
\end{figure}

The effects of public transportation structure types on crime are shown in Figures 
\ref{fig:mtl_pubtransport} and \ref{fig:mt_pubtransport}. As was the case for abandoned buildings, while the effect of public transportation infrastructure is observable at a regional level, Figure \ref{fig:mt_pubtransport} shows that much of it is revealed to be highly localized: crime density is elevated close to public transportation infrastructure and decreases rapidly with distance toward a flat asymptote that is similar to the control series. We hypothesize that this localization results from certain characteristics of foot traffic induced by public transportation infrastructure. It is likely that this traffic fluctuates according to a highly consistent temporal pattern (i.e., rush hour), and the individuals comprising this traffic likely follow the same routes every day. This patterning may incentivize crime around stations, given how predictable the movement of potential victims is. 

The effects of public transportation infrastructure are also inconsistent between cities. Rail stations and subway stations are not directly comparable because the former are mostly aboveground while the latter are mostly below ground, so we focus on differences in the effects of bus stops. In Chicago, bus stops are associated with a greater regional effect, and density analysis indicates that their effects are less localized. In comparison, bus stops in New York City exert a less pronounced regional effect, but density analysis reveals an extremely localized impact on crime. 

\begin{figure}[h]
    \centering
    \begin{subfigure}[b]{0.48\textwidth}
        \centering
        \includegraphics[width=\textwidth]{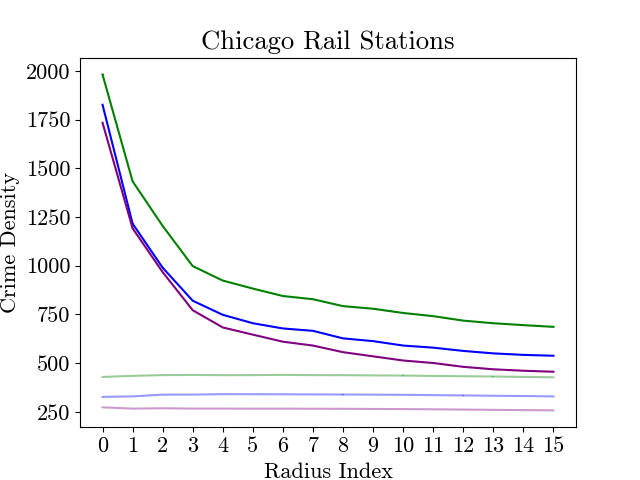}
        \label{fig:mt_chicago_rail_stations}
    \end{subfigure}
    \begin{subfigure}[b]{0.48\textwidth}
        \centering
        \includegraphics[width=\textwidth]{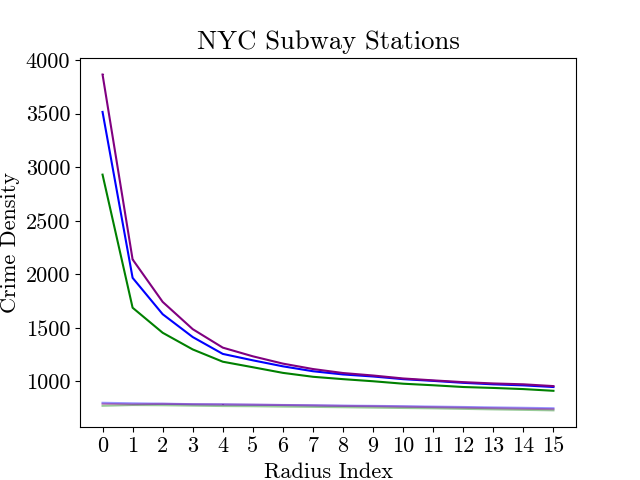}
        \label{fig:mt_nyc_subway_stations}
    \end{subfigure}
\end{figure}
\begin{figure}\ContinuedFloat
    \begin{subfigure}[b]{0.48\textwidth}
        \centering
        \includegraphics[width=\textwidth]{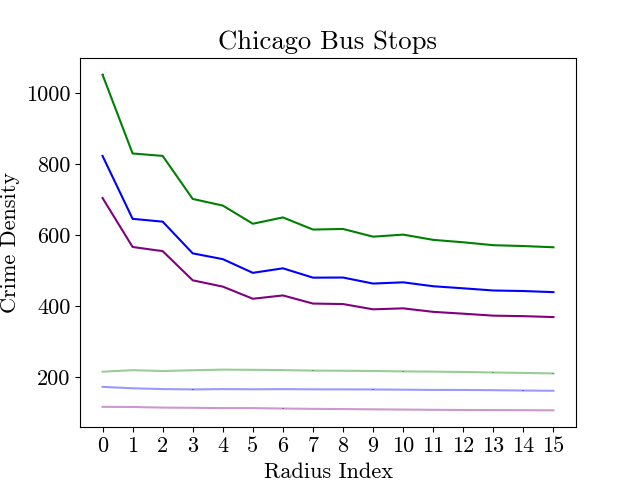}
        \label{fig:mt_chicago_bus_stops}
    \end{subfigure}
    \begin{subfigure}[b]{0.48\textwidth}
        \centering
        \includegraphics[width=\textwidth]{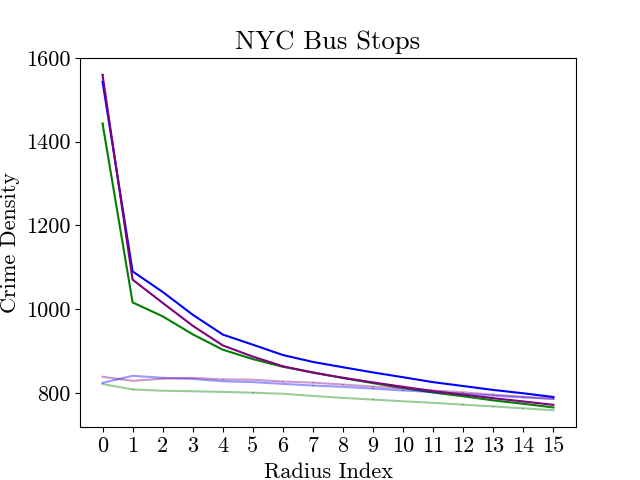}
        \label{fig:mt_nyc_bus_stops}
    \end{subfigure}
    \begin{subfigure}[b]{0.6\textwidth}
        \centering
        \fbox{\includegraphics[width=\textwidth]{figs/legend.png}}
    \end{subfigure}
    \vspace{0.1in}
    \caption{ \textit{Density analysis shows the effects of public transportation infrastructure structure types are highly localized}. In both Chicago and New York, the crime density is elevated near rail stations and subway stations, respectively, much higher than the controls, and drops as we move farther away from the structure types of interest. This pattern is more pronounced in New York, where proximity to bus stops results in a larger treatment effect compared to Chicago. However, in NYC bus stops, crime density overlaps with treated tracts after getting far enough away from a bus stop, indicating a very localized effect of bus stops in NYC.}
    \label{fig:mt_pubtransport}
\end{figure}

\begin{figure}[h]
    \centering
    \begin{subfigure}[b]{0.48\textwidth}
        \centering
        \includegraphics[width=\textwidth]{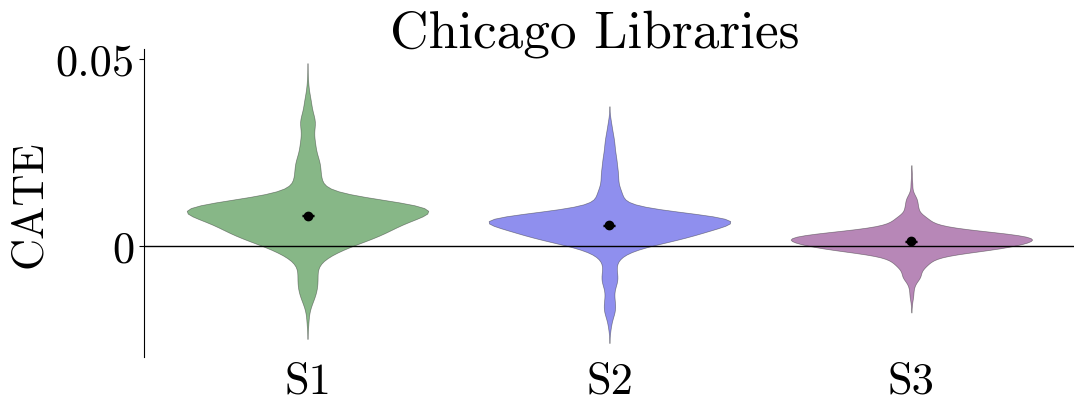}
        \label{fig:mtl_chicago_libraries}
    \end{subfigure}
    \hfill
    \begin{subfigure}[b]{0.48\textwidth}
        \centering
        \includegraphics[width=\textwidth]{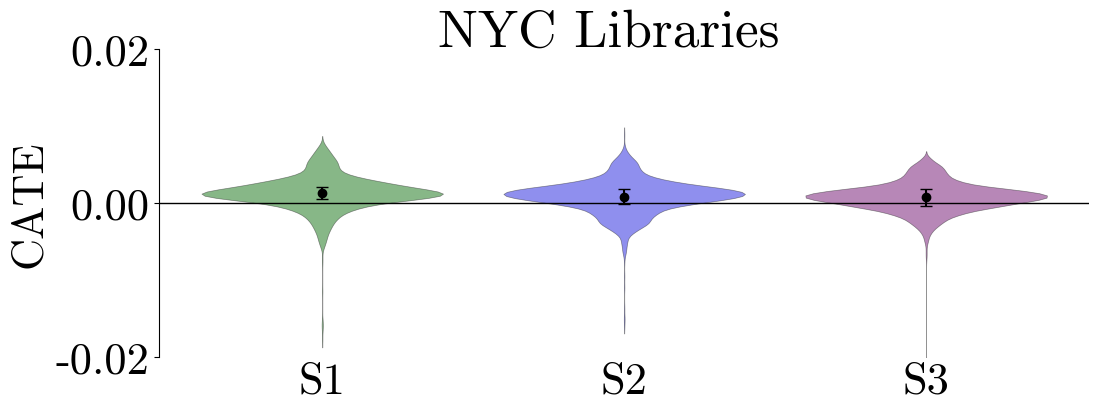}
        \label{fig:mtl_nyc_libraries}
    \end{subfigure}
    \begin{subfigure}[b]{0.48\textwidth}
        \centering
        \includegraphics[width=\textwidth]{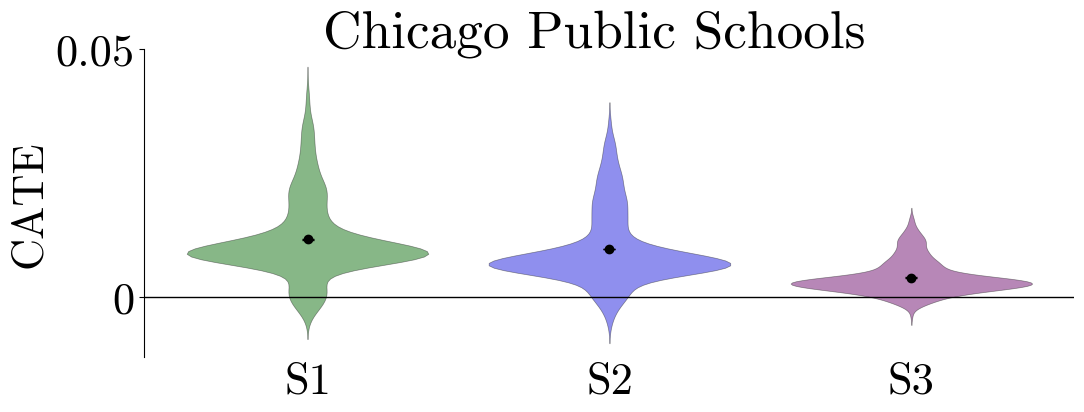}
        \label{fig:mtl_chicago_public_schools}
    \end{subfigure}
    \hfill
    \begin{subfigure}[b]{0.48\textwidth}
        \centering
        \includegraphics[width=\textwidth]{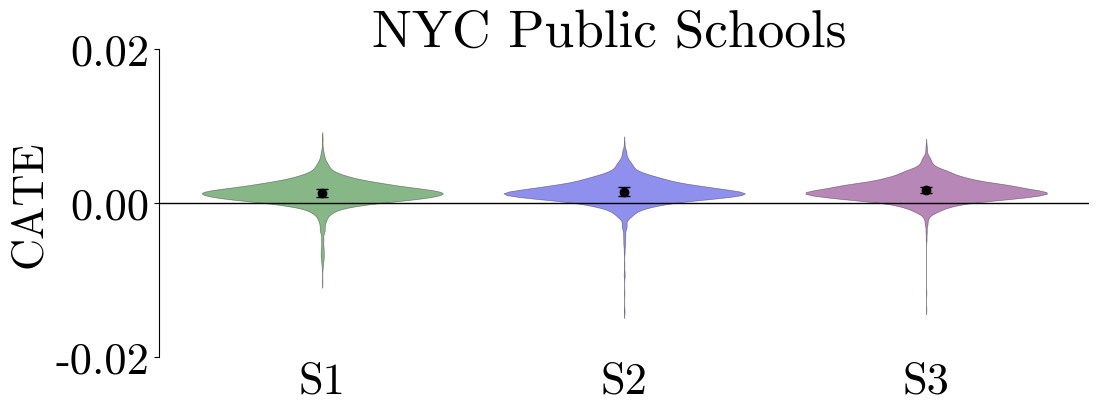}
        \label{fig:mtl_nyc_public_schools}
    \end{subfigure}
    \caption{\textit{Tracts with libraries and public schools have higher crime rates than those without}. While both structure types appear to induce higher tract-level crime rates, with positive CATE distributions, they differ between the cities, with the structure types having a higher impact in Chicago. }
    \label{fig:mtl_pslib}
\end{figure}

\begin{figure}[h]
    \centering
    \begin{subfigure}[b]{0.48\textwidth}
        \centering
        \includegraphics[width=\textwidth]{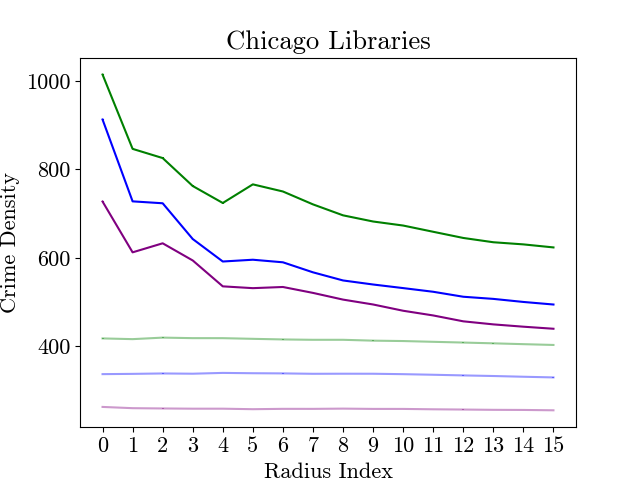}
        \label{fig:mt_chicago_libraries}
    \end{subfigure}
    \begin{subfigure}[b]{0.48\textwidth}
        \centering
        \includegraphics[width=\textwidth]{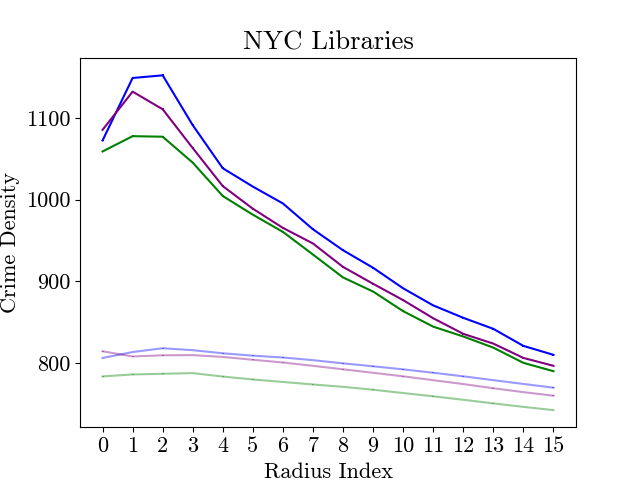}
        \label{fig:mt_nyc_libraries}
    \end{subfigure}
    \begin{subfigure}[b]{0.48\textwidth}
        \centering
        \includegraphics[width=\textwidth]{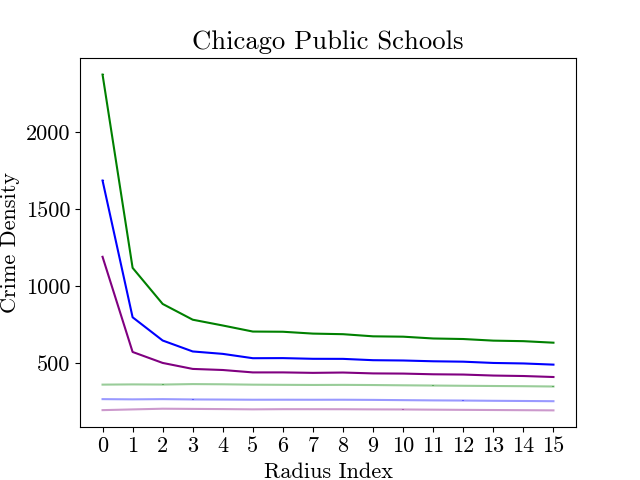}
        \label{fig:mt_chicago_public_schools}
    \end{subfigure}
    \begin{subfigure}[b]{0.48\textwidth}
        \centering
        \includegraphics[width=\textwidth]{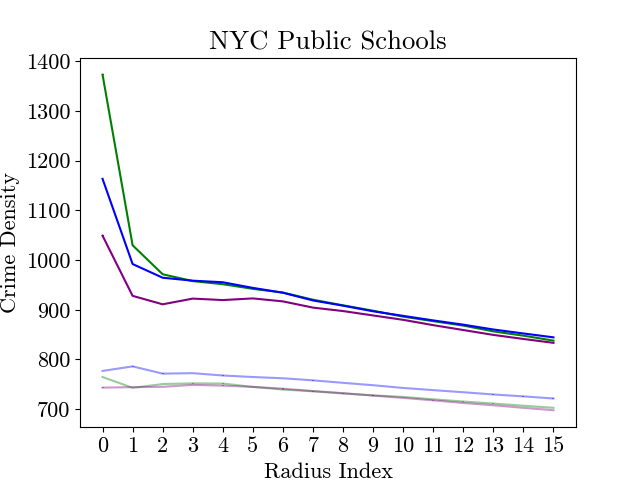}
        \label{fig:mt_nyc_public_schools}
    \end{subfigure}
    \vspace{0.1in}
    \begin{subfigure}[b]{0.6\textwidth}
        \centering
        \fbox{\includegraphics[width=\textwidth]{figs/legend.png}}
    \end{subfigure}
    \vspace{0.1in}
    \caption{\textit{ Density analysis shows the libraries and public schools have strong local effects on crime and differ between cities}. In Chicago, the effects of both libraries and public schools decrease with distance, as evident by the smaller difference between crime densities between treatment and control tracts. In New York, the effect of libraries briefly increases with distance, before then decreasing until hitting an asymptote above the control.}
    \label{fig:mt_pslib}
\end{figure}

Figures \ref{fig:mtl_pslib} and \ref{fig:mt_pslib} depict the effects of libraries and public schools. In contrast to public transportation infrastructure, the foot traffic drawn by these structure types is likely comprised of individuals who live nearby. However, this traffic probably follows a similar pattern where certain times of day are busier. This is especially true for public schools.

Regional effects of both structure types are greater in Chicago, which continues to experience reduced effects in S3. For libraries in particular, both cities have periods in which libraries do not have observable effects on tract-level crime. However, both structure types have large effects on more localized areas. We note that the effect of libraries in New York is unlike that of other structure types given that it increases first before decreasing linearly with distance. 


\begin{figure}[h]
    \centering
    \begin{subfigure}[b]{0.48\textwidth}
        \centering
        \includegraphics[width=\textwidth]{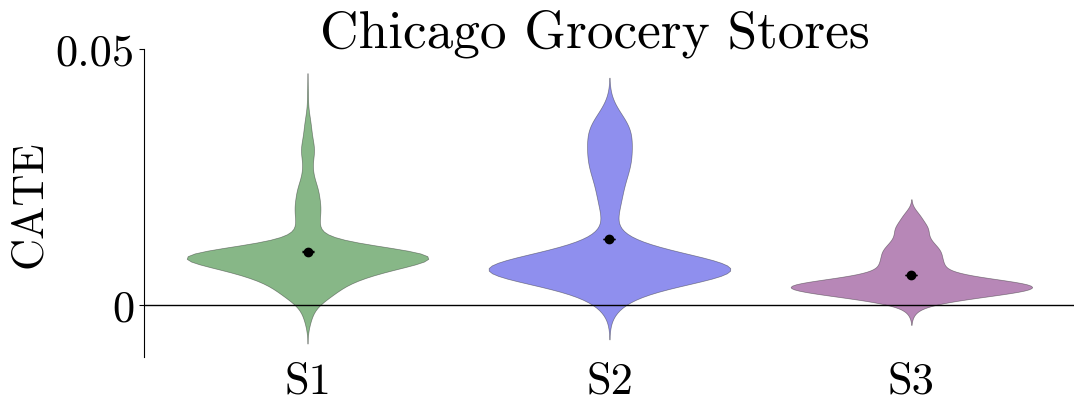}
        \label{fig:mtl_chicago_grocery_stores}
    \end{subfigure}
    \hfill
    \begin{subfigure}[b]{0.48\textwidth}
        \centering
        \includegraphics[width=\textwidth]{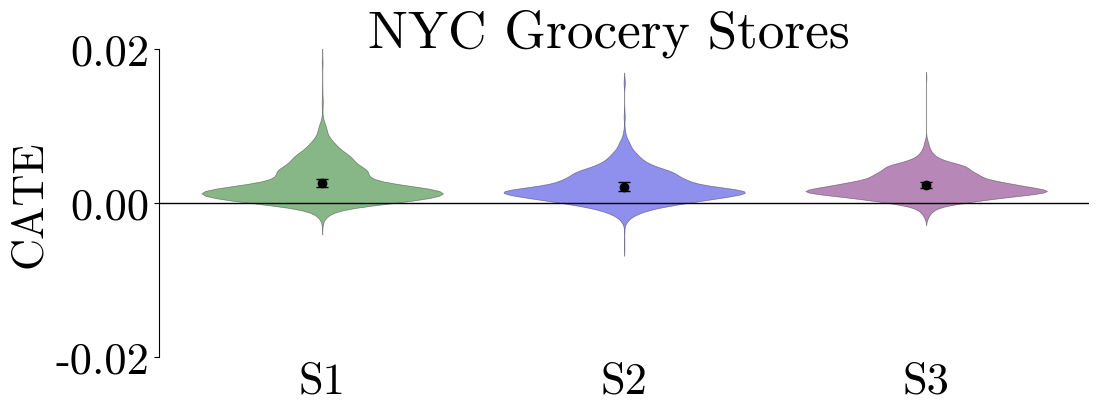}
        \label{fig:mtl_nyc_grocery_stores}
    \end{subfigure}
    \\
    \begin{subfigure}[b]{0.48\textwidth}
        \centering
        \includegraphics[width=\textwidth]{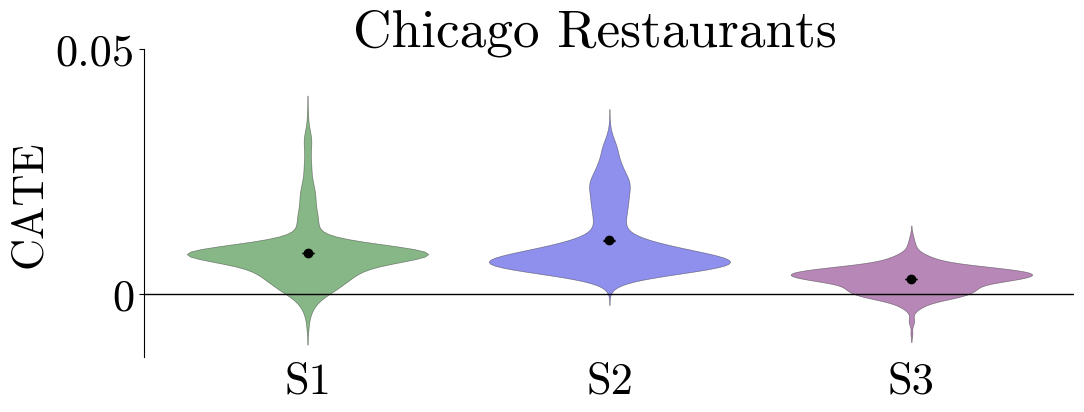}
        \label{fig:mtl_chicago_restaurants}
    \end{subfigure}
    \hfill
    \begin{subfigure}[b]{0.48\textwidth}
        \centering
        \includegraphics[width=\textwidth]{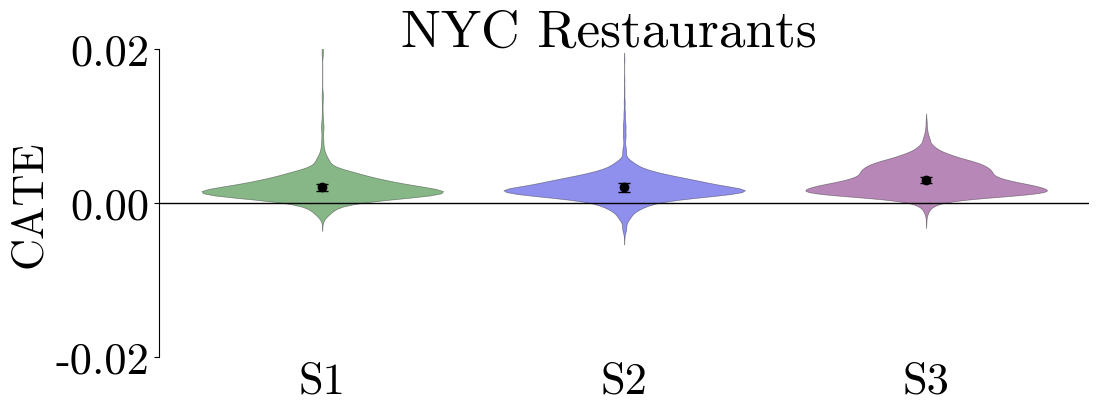}
        \label{fig:mtl_nyc_restaurants}
    \end{subfigure}
    \caption{\textit{Tracts with a large number of grocery stores and restaurants have higher crime rates}. Similar to libraries and public schools, the effect of these structure types continues to be greater in Chicago and continues to decrease in S3.}
    \label{fig:mtl_gsr}
\end{figure}

\begin{figure}[h]
    \centering
    \begin{subfigure}[b]{0.48\textwidth}
        \centering
        \includegraphics[width=\textwidth]{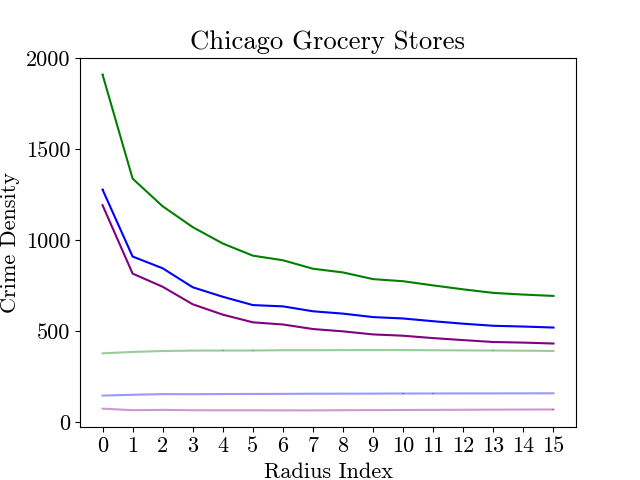}
        \label{fig:mt_chicago_grocery_stores}
    \end{subfigure}
    \begin{subfigure}[b]{0.48\textwidth}
        \centering
        \includegraphics[width=\textwidth]{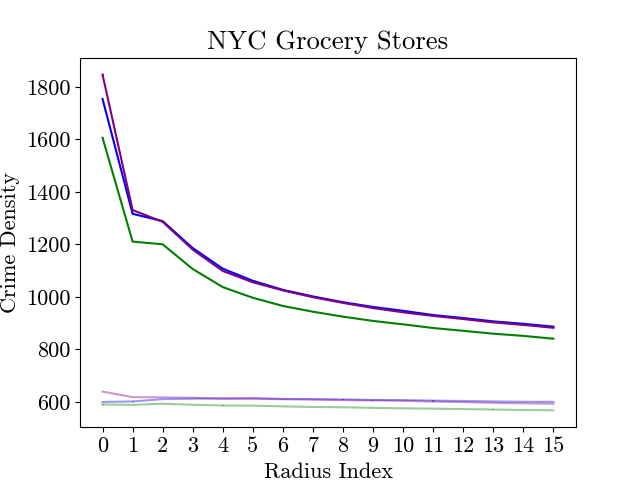}
        \label{fig:mt_nyc_grocery_stores}
    \end{subfigure}
    \\
    \begin{subfigure}[b]{0.48\textwidth}
        \centering
        \includegraphics[width=\textwidth]{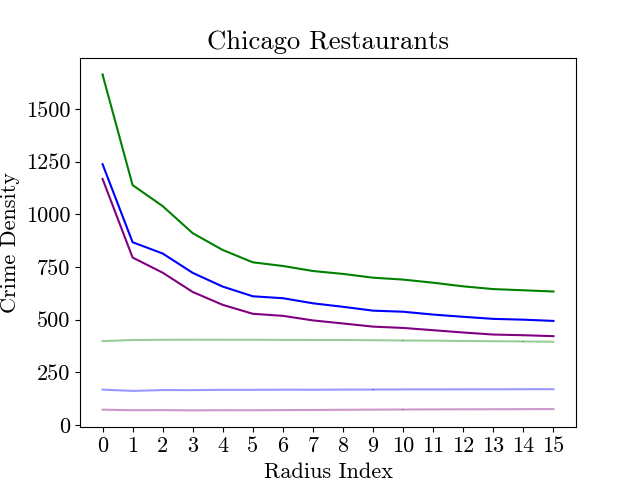}
        \label{fig:mt_chicago_restaurants}
    \end{subfigure}
    \begin{subfigure}[b]{0.48\textwidth}
        \centering
        \includegraphics[width=\textwidth]{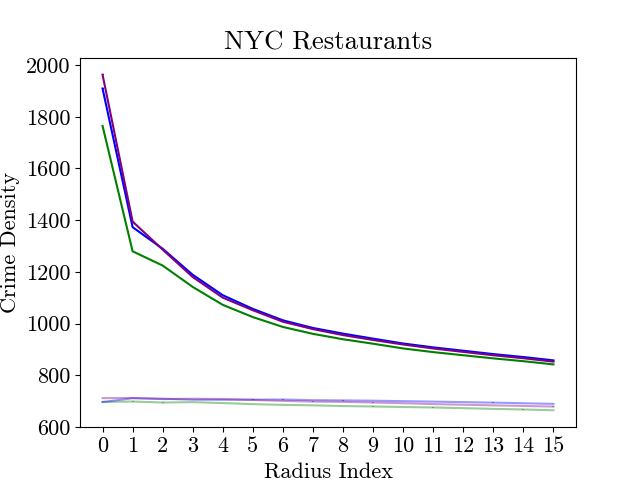}
        \label{fig:mt_nyc_restaurants}
    \end{subfigure}
    \vspace{0.1in}
    \begin{subfigure}[b]{0.6\textwidth}
        \centering
        \fbox{\includegraphics[width=\textwidth]{figs/legend.png}}
    \end{subfigure}
    \vspace{0.1in}
    \caption{\textit{Treatment effects in grocery stores and restaurants are localized but still affect tracts}. In these structure types, the crime density is elevated while close but then tapers off to an asymptote above the control series. This indicates that, while they still have tract-level effects, they have very localized effects as well. 
    }
    \label{fig:mt_gsr}
\end{figure}

Figures \ref{fig:mtl_gsr} and \ref{fig:mt_gsr} depict the effects of grocery stores and restaurants. Similar to previously analyzed structure types, we hypothesize that the foot traffic generated by stores and restaurants likely follows a chronological pattern, which may incentivize crime due to the regularity of movements. The same general patterns continue to hold. At the tract level, these structure types have more of an effect in Chicago which decreases in S3. These effects are also highly localized. 


\subsection{BWT for Perceived Danger}
\label{subsec:perceivedres}

\begin{figure}[h]
    \centering
    \includegraphics[width=0.8\linewidth]{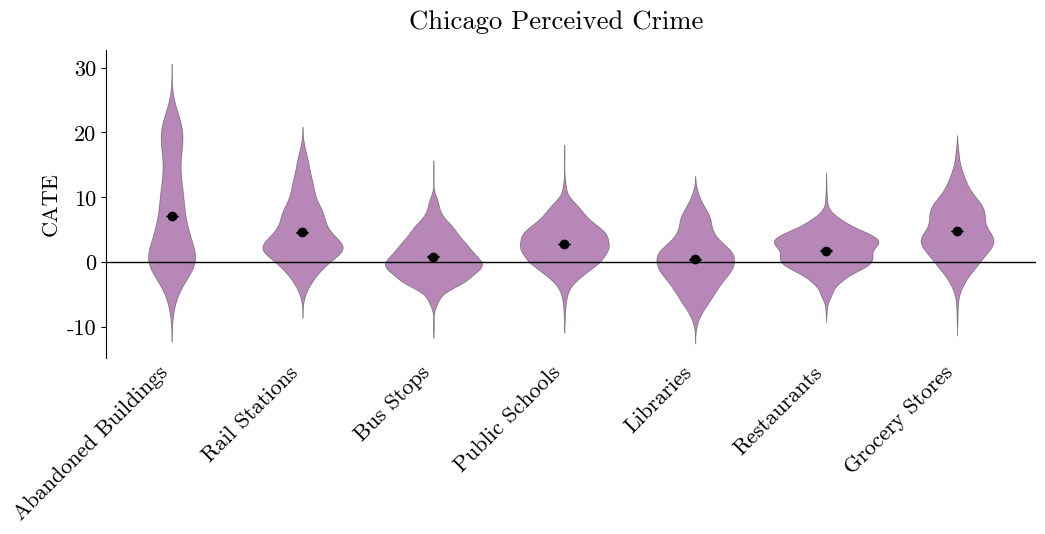}
    \caption{Effect of previously analyzed structures on \emph{perceived} crime in Chicago during 2021 and 2022. \textit{BWT extends to perceived crime: abandoned buildings, along with other structure types related to heavy foot traffic (except libraries), all have a large positive effect on perceived crime.}}
    \label{fig:mtl_chicago_perceived_crime}
\end{figure}

Figure \ref{fig:mtl_chicago_perceived_crime} depicts the effects of previously discussed structure types on perceived danger in Chicago. The large positive effect of abandoned buildings indicates that the broken windows effect extends to perceptions of safety. This trend extends to types of structures that draw heavy foot traffic: all structure types except libraries (which did not have a large effect on regional crime in Chicago during S3) exert positive effects on perceived crime rate. This seems to support a core premise of BWT: that perceived danger is linked to actual danger, though these results alone are insufficient to establish any direction in causality. 

We note that our metrics for tract-level crime rate and perceived danger are not directly comparable. Instead, we compare the ordinal ranking of structure types by their estimated ATEs on real and perceived danger in Table \ref{table:pvsa}. For both outcomes, the top three and bottom four ranked structure types are the same, implying that perceived and actual crime may be more correlated than comparing treatment effect significance would suggest.

\begin{table}[h]
\begin{tabular}{c|c}
{\textbf{Ranked by ATE on Crime Rate}} & {\textbf{Ranked by ATE on Perceived Danger}} \\ \hline
\begin{tabular}{c}
     Abandoned Buildings \\
     $ATE = 8.57 \times 10^{-3} \pm 6.00 \times 10^{-5}$
\end{tabular}
& 
\begin{tabular}{c}
Abandoned Buildings \\
$ATE = 7.12 \pm 5.66 \times 10^{-2}$
\end{tabular} 
\\
\begin{tabular}{c}
Rail Stations \\
$ATE = 7.18 \times 10^{-3} \pm 8.10 \times 10^{-5}$
\end{tabular} 
&
\begin{tabular}{c}
Grocery Stores \\
$ATE = 4.78 \pm 5.00 \times 10^{-2}$
\end{tabular} 
\\
\begin{tabular}{c}
Grocery Stores \\
$ATE = 6.32 \times 10^{-3} \pm 5.50 \times 10^{-5}$
\end{tabular} 
&
\begin{tabular}{c}
Rail Stations \\
$ATE = 4.59 \pm 7.27 \times 10^{-2}$
\end{tabular} 
\\
\begin{tabular}{c}
Bus Stops \\
$ATE = 4.29 \times 10^{-3} \pm 5.60 \times 10^{-5}$
\end{tabular} 
&
\begin{tabular}{c}
Public Schools \\
$ATE = 2.83 \pm 5.01 \times 10^{-2}$
\end{tabular} 
\\
\begin{tabular}{c}
Public Schools \\
$ATE = 3.88 \times 10^{-3} \pm 5.50 \times 10^{-5}$
\end{tabular} 
&
\begin{tabular}{c}
Restaurants \\
$ATE = 1.77 \pm 5.07 \times 10^{-2}$
\end{tabular} 
\\
\begin{tabular}{c}
Restaurants \\
$ATE = 2.94 \times 10^{-3} \pm 5.70 \times 10^{-5}$
\end{tabular} 
&
\begin{tabular}{c}
Bus Stops \\
$ATE = 8.32 \times 10^{-1} \pm 5.04 \times 10^{-2}$
\end{tabular} 
\\
\begin{tabular}{c}
Libraries \\
$ATE = 1.30 \times 10^{-4} \pm 8.80 \times 10^{-5}$
\end{tabular} 
& 
\begin{tabular}{c}
Libraries \\
$ATE = 4.40 \times 10^{-1} \pm 7.73 \times 10^{-2}$
\end{tabular}
\end{tabular} 
\vspace{0.1in}
\caption{\textit{Ordinal rankings of structures by estimated ATE on crime and perceived danger.} Actual crime rate and perceived danger are correlated, as the top three and bottom four ranked structure types are the same.}
\label{table:pvsa}
\end{table}

\subsection{Treatment Effect Heterogeneity}
\label{subsec:heterogeneity}

We report CATE estimates for selected treatments over several demographic subgroups that are particularly informative for highlighting heterogeneity (or lack thereof) in treatment outcomes, while omitting cases where results are uninformative and redundant for the purpose of exposition. This subgroup analysis helps reveal whether and how different communities experience the effects of the built environment differently.

\begin{figure}[h]
    \centering
    \begin{subfigure}[b]{0.9\textwidth}
        \centering
        \includegraphics[width=0.8\textwidth]{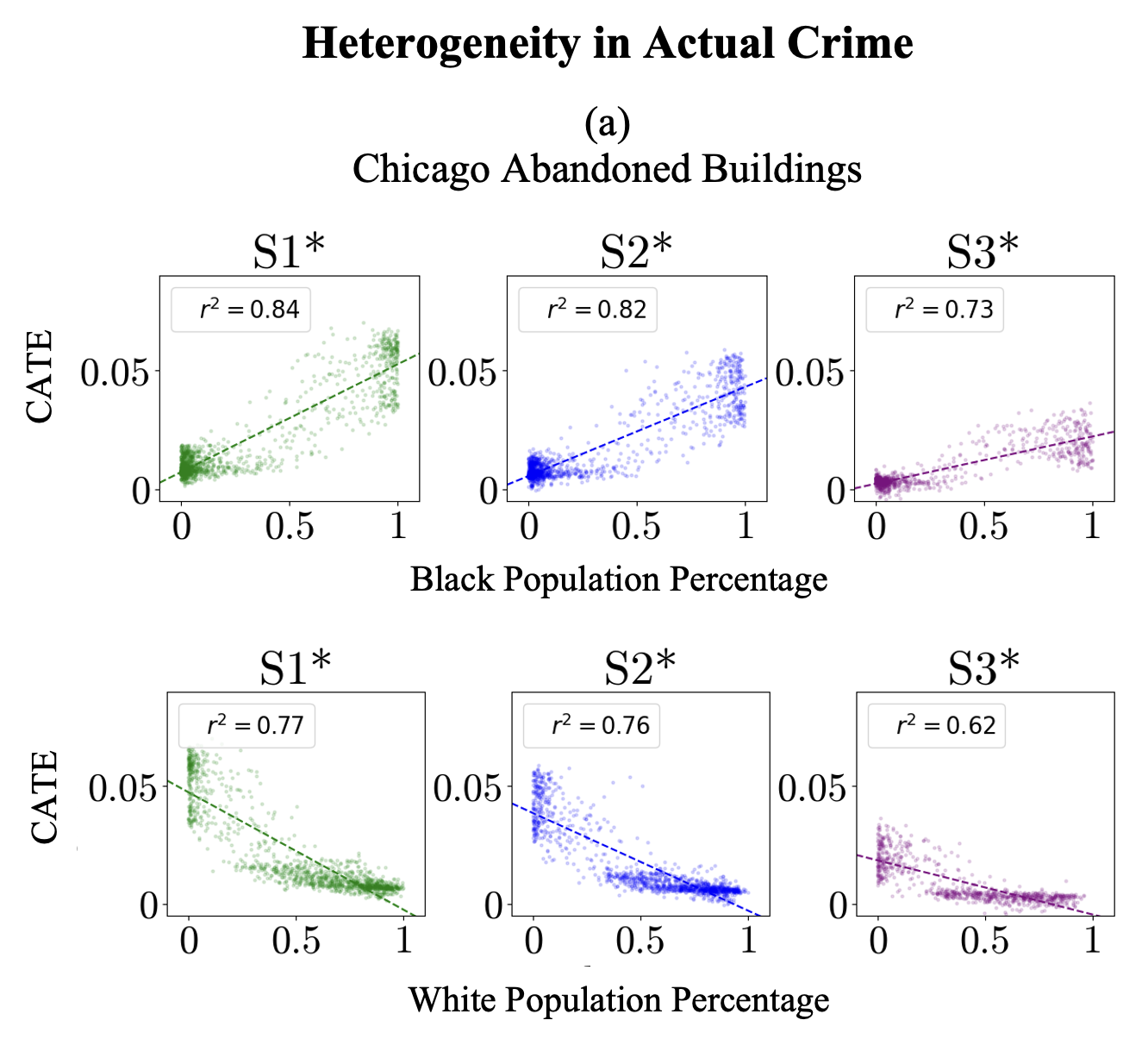}
        \label{fig:mtl_chicago_abandoned_buildings_black_pop_scatter_actual_crime}
    \end{subfigure}
    \hfill
    \begin{subfigure}[b]{0.9\textwidth}
        \centering
        \includegraphics[width=0.8\textwidth]{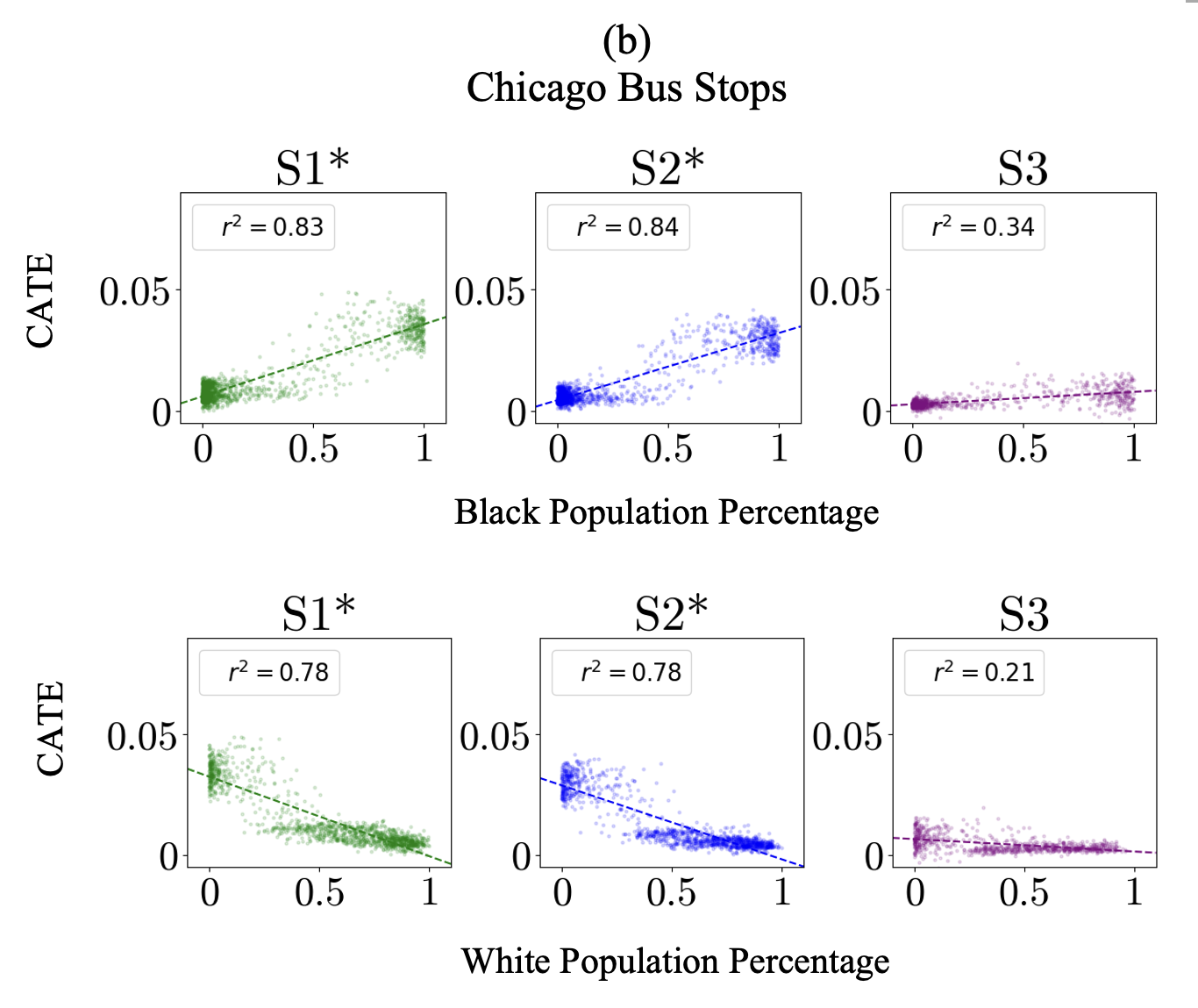}
        \label{fig:mtl_chicago_bus_stops_white_pop_scatter_actual_crime}
    \end{subfigure}
    \caption{\textit{Treatment effect heterogeneity on actual crime in Chicago, according to racial subgroups.} Heterogeneity in actual crime across subgroups defined by racial composition was especially common, particularly for Black and White populations. The proportion of Black populations within a tract was correlated positively with the treatment effect of all five structure types considered here, while the proportion of white people was correlated negatively. Starred plots represent substantial heterogeneity.}
    \label{fig:heterogeneity-race-actual}
\end{figure}

\begin{figure}[h]
    \centering
    \begin{subfigure}[b]{\textwidth}
        \centering
        \includegraphics[width=0.8\textwidth]{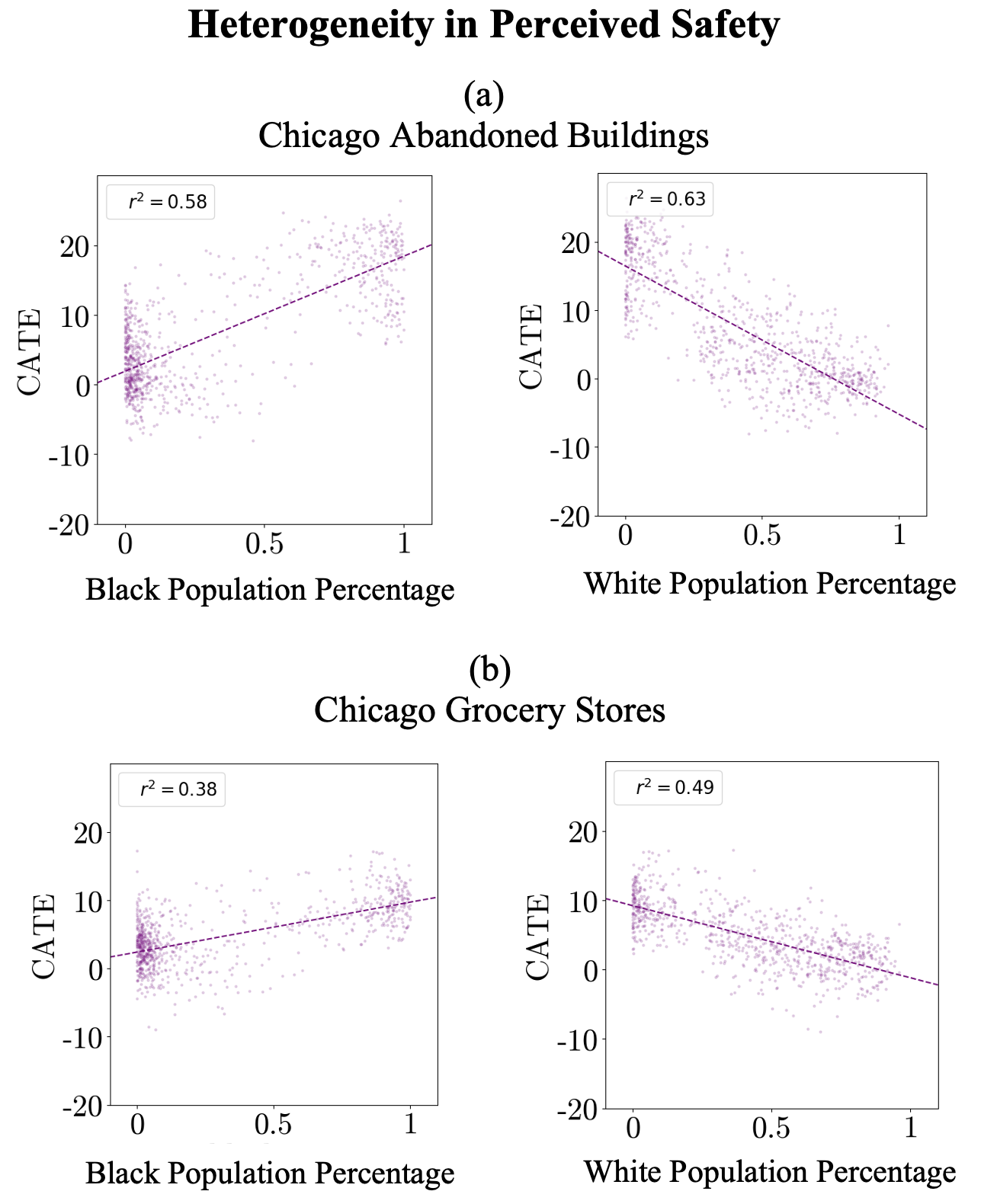}
        \label{fig:mtl_chicago_abandoned_buildings_black_pop_scatter_perceived_crime}
    \end{subfigure}
    \caption{\textit{Treatment effect heterogeneity on perceived safety in Chicago, according to racial subgroups.} Heterogeneity across racial composition was also observed for perceived safety.}
    \label{fig:chi-heterogeneity-race-perceived}
\end{figure}

\begin{figure}[h]
    \centering
    \begin{subfigure}[b]{0.9\textwidth}
        \centering
        \includegraphics[width=\textwidth]{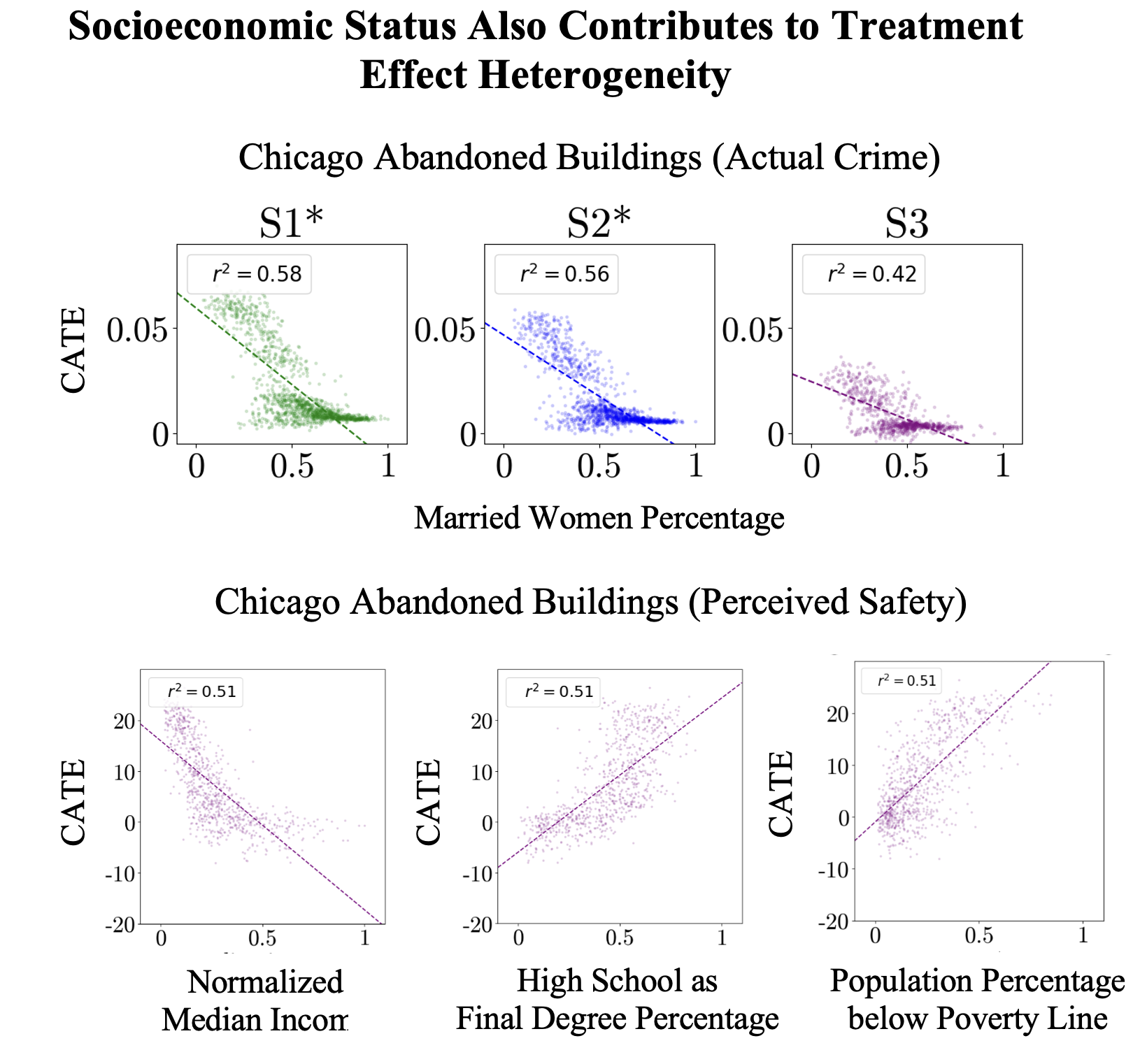}
        \label{fig:mtl_chicago_abandoned_buildings_female_married_scatter_actual_crime}
    \end{subfigure}
    \caption{\textit{Treatment effect heterogeneity on actual and perceived crime in Chicago, according to socioeconomic subgroups.} Socioeconomic status was another major contributor to treatment effect heterogeneity. In addition to married women, median income, high school as the final degree, and population below the poverty line, variables not shown here include unemployment rates and married men.}
    \label{fig:chi-heterogeneity-ses}
\end{figure}

In Chicago, there was substantial treatment effect heterogeneity on crime ($r^2 \geq 0.5$) over at least one subgroup for five structure types -- all except restaurants and libraries. The effects of grocery stores and abandoned buildings on perceived safety were also heterogeneous. The most common subgroups for which heterogeneity was documented were racial (Figures \ref{fig:heterogeneity-race-actual} and \ref{fig:chi-heterogeneity-race-perceived}) or socioeconomic, with a notable exception being marital status (Figure \ref{fig:chi-heterogeneity-ses}).

\begin{figure}[h]
    \centering
    \begin{subfigure}[b]{0.9\textwidth}
        \centering
        \includegraphics[width=\textwidth]{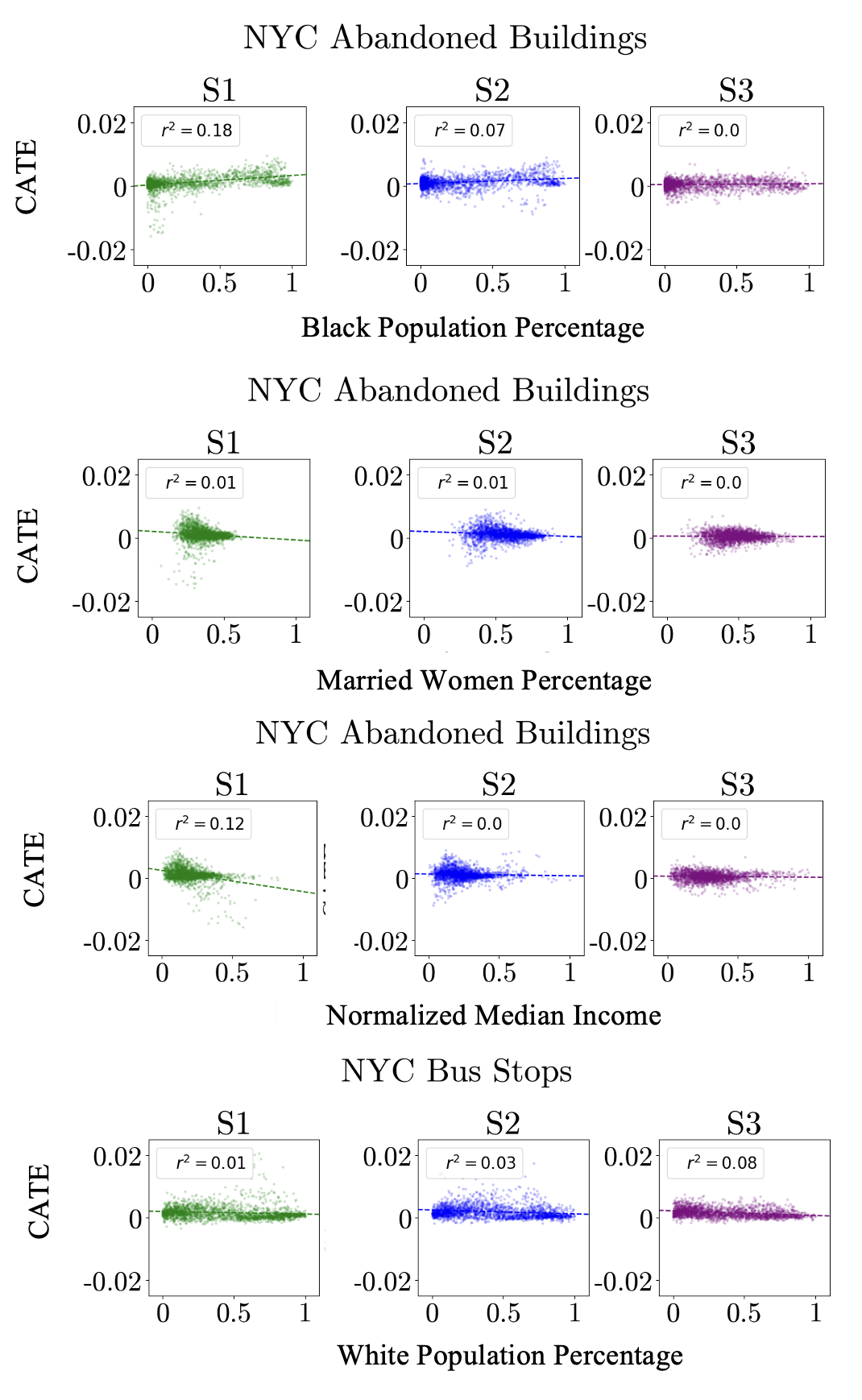}
        \label{fig:mtl_nyc_abandoned_buildings_black_pop_scatter}
    \end{subfigure}
    \caption{\textit{Treatment effect heterogeneity on actual crime in New York City.} Structure types in New York City did not experience treatment effect heterogeneity for any subgroup.}
    \label{fig:nyc-heterogeneity}
\end{figure}

In contrast, there was almost no heterogeneity in New York at the same significance level. For comparison, we include results from New York for several of the structure/covariate pairs described above in Figure \ref{fig:nyc-heterogeneity}. 

These patterns have several important implications. First, they highlight the need for fine-grained and disaggregated impact analysis when designing urban interventions. Second, the heterogeneity across subgroups suggests that community and localized planning may be crucial for maximizing the effectiveness of crime intervention strategies. For example, interventions like converting abandoned structures into green space may reduce crime in tracts with a certain socioeconomic demographic composition but have a less pronounced impact on others. For future work, it would be interesting to explore and explain the underlying mechanisms driving these heterogeneous responses, which may allow for a clearer picture of crime intervention strategies.

\section{Conclusion}
\label{sec:conclusion}

The goal of this study was to leverage observational data to perform a causal analysis on the effect of urban design on crime and public perception of danger. To do so, we applied the MALTS matching algorithm to estimate the effect of various urban structures on crime at different spatial resolutions in Chicago and New York City as well as to estimate the effect of these structures on perceived danger in Chicago. The direction of this study was guided primarily by BWT and by Jane Jacobs' ``eyes on the street'' hypothesis. Our aim was specifically to re-evaluate past results with the benefit of larger, more detailed datasets and more accurate and auditable methods of treatment effect estimation, and to address shortcomings in existing literature pertaining to the psychological impact of urban design and the effects of foot traffic. 

Our results on the validity of BWT and Jane Jacobs' theory are thus mixed; although we do find that signs of blight (e.g., abandoned buildings) are indeed criminogenic, we also find that signs of density and vibrancy are associated both with more crime and increased feelings of danger. One potential shortcoming of our study is that we include physical structures related to density but compute crime rates according to resident population (i.e., the number of people who live in a given area), rather than ambulatory population (the number of people who traffic through the same area), but our findings square with those of \citet{nadai2020socio}, who explicitly consider the latter. Their results are also consistent with ours in favoring Oscar Newman's prediction that density is associated with greater levels of crime.

Our approaches highlight that the effects of urban design on both real and perceived danger are location-dependent. This complements previous multi-city studies \citep[e.g.,][]{nadai2020socio, barnum2017kaleidoscope} which have similarly found heterogeneous treatment effects at the city level. From a theoretical perspective, this suggests that one-size-fits-all approaches to understanding crime are insufficient. Even within cities, our results also highlight the importance of treatment effect heterogeneity, which must be accounted for in understanding how changes in the urban environment will influence crime, both real and perceived. These observational results should inform analysis of urban crime at a greater degree of granularity, rather than the broad resolution typically adopted in studies. 

From a policy perspective, our results on perceived danger add to a growing literature documenting the link between subjective beliefs about public safety and determinants of actual threat risk. We argue that both actual safety and perceived danger are public goods worthy of policy attention.

\subsection*{Disclosure Statement}
The authors have no conflicts of interest to declare.

\subsection*{Acknowledgments}
The authors thank the reviewers for their commentary and feedback. 
 
\subsection*{Contributions}
ZC, EJ, NS, and HW contributed equally to the research endeavor, from data gathering and cleaning to coding and drafting. EC and CR provided plentiful guidance and advice, without which this project would not have been realized.

\appendix

\section{Demographic Variables}
\label{appendix:demographic variables}

The 30 demographic variables that were drawn from the ACS five-year estimates and used in our analysis are listed below.

\subsection*{Population Size and Sex Ratio (Three Features)}
\begin{enumerate}
    \item \textbf{total\_pop} - Total Population  
    \item \textbf{male\_pop} - Male Population  
    \item \textbf{female\_pop} - Female Population  
\end{enumerate}

\subsection*{Age (Six Features)}
\begin{enumerate}
    \item \textbf{m\_juv} - Males 0-19 Years  
    \item \textbf{m\_ad} - Males 21 to 65 Years  
    \item \textbf{m\_eld} - Males over 65 Years  
    \item \textbf{f\_juv} - Females 0-19 Years  
    \item \textbf{f\_ad} - Females 21 to 65 Years  
    \item \textbf{f\_eld} - Females over 65 Years  
\end{enumerate}

\subsection*{Household Structure (Seven Features)}
\begin{enumerate}
    \item \textbf{num\_households} - Number of Households  
    \item \textbf{male\_never\_married} - Males Never Married  
    \item \textbf{male\_married} - Married Males  
    \item \textbf{male\_divorced} - Divorced Males  
    \item \textbf{female\_never\_married} - Females Never Married  
    \item \textbf{female\_married} - Married Females  
    \item \textbf{female\_divorced} - Divorced Females  
\end{enumerate}

\subsection*{Race (Six Features)}
\begin{enumerate}
    \item \textbf{white\_pop} - White Population  
    \item \textbf{black\_pop} - Black Population  
    \item \textbf{nat\_pop} - Native American Population  
    \item \textbf{asian\_pop} - Asian Population  
    \item \textbf{mixed\_pop} - Mixed Race Population  
    \item \textbf{other\_pop} - Other Race Population  
\end{enumerate}

\subsection*{Educational Attainment (Four Features)}
\begin{enumerate}
    \item \textbf{no\_highschool} - Population Without High School Diploma  
    \item \textbf{highschool} - Population With High School Diploma as their Final Degree
    \item \textbf{undergrad} - Population With Undergraduate Degree as their Final Degree
    \item \textbf{postgrad} - Population With Postgraduate Degree as their Final Degree
\end{enumerate}

\subsection*{Socioeconomic Status (Two Features)}
\begin{enumerate}
    \item \textbf{median\_income} - Median Income  
    \item \textbf{poverty} - Population Below Poverty Line  
\end{enumerate}

\subsection*{Labor Force Statistics (Two Features)}
\begin{enumerate}
    \item \textbf{in\_labor} - Population in Labor Force  
    \item \textbf{unemployed} - Unemployed Population  
\end{enumerate}

\FloatBarrier
\newpage
\printbibliography

@article{wilson1982broken,
    title={Broken Windows: The police and neighborhood safety},
    author={Wilson, James Q and Kelling, George L},
    journal= {{Atlantic Monthly}},
    volume={249},
    number={3},
    pages={29--38},
    year={1982},
    url = {https://www.theatlantic.com/magazine/archive/1982/03/broken-windows/304465/}
}

@article{macdonald2021reducing,
    title={Reducing crime by remediating vacant lots: the moderating effect of nearby land uses},
    author={MacDonald, John M and Nguyen, Viet and Jensen, Shane T and Branas, Charles C},
    journal= {{Journal of Experimental Criminology}},
    pages={1--26},
    year={2021},
    publisher={Springer},
    doi = {10.1007/s11292-020-09452-9}
}

@article{burgason2017close,
    title={Close only counts in alcohol and violence: Controlling violence near late-night alcohol establishments using a routine activities approach},
    author={Burgason, Kyle A and Drawve, Grant and Brown, Timothy C and Eassey, John},
    journal= {{Journal of Criminal Justice}},
    volume={50},
    pages={62--68},
    year={2017},
    publisher={Elsevier},
    doi = {10.1016/j.jcrimjus.2017.04.004}
}

@article{groff2012role,
    title={The role of neighborhood parks as crime generators},
    author={Groff, Elizabeth and McCord, Eric S},
    journal= {{Security Journal}},
    volume={25},
    pages={1--24},
    year={2012},
    publisher={Springer},
    doi = {10.1057/sj.2011.1}
}

@article{connealy2020can,
    title={Can we trust crime predictors and crime categories?},
    subtitle = {Expansions on the potential problem of generalization},
    author={Connealy, Nathan T},
    journal= {{Applied Spatial Analysis and Policy}},
    volume={13},
    pages={669--692},
    year={2020},
    publisher={Springer},
    doi = {10.1007/s12061-019-09323-5}
}

@article{WangEtAlFLAME2021,
  author  = {Tianyu Wang and Marco Morucci and M. Usaid Awan and Yameng Liu and Sudeepa Roy and Cynthia Rudin and Alexander Volfovsky},
  title   = {FLAME: A Fast Large-scale Almost Matching Exactly Approach to Causal Inference},
  journal = {Journal of Machine Learning Research},
  year    = {2021},
  volume  = {22},
  number  = {31},
  pages   = {1-41},
  url     = {http://jmlr.org/papers/v22/19-853.html}
}

@article{parikh2022malts,
    title={{MALTS}: Matching after learning to stretch},
    author={Parikh, Harsh and Rudin, Cynthia and Volfovsky, Alexander},
    journal= {{Journal of Machine Learning Research}},
    volume={23},
    number={240},
    pages={1--42},
    year={2022},
    url = {http://jmlr.org/papers/v23/21-0053.html}
}

@article{entorf2000socioeconomic,
    title={Socioeconomic and demographic factors of crime in {G}ermany: Evidence from panel data of the {G}erman states},
    author={Entorf, Horst and Spengler, Hannes},
    journal= {{International Review of Law and Economics}},
    volume={20},
    number={1},
    pages={75--106},
    year={2000},
    publisher={Elsevier},
    doi = {10.1016/S0144-8188(00)00022-3}
}

@article{stucky2016intra,
    title={Intra-and inter-neighborhood income inequality and crime},
    author={Stucky, Thomas D and Payton, Seth B and Ottensmann, John R},
    journal= {{Journal of Crime and Justice}},
    volume={39},
    number={3},
    pages={345--362},
    year={2016},
    publisher={Taylor \& Francis},
    doi = {10.1080/0735648X.2015.1004551}
}

@techreport{pewlocalcrime,
  title = {Americans' Experience With Local Crime News},
  author = {Eddy, Kirsten and Lipka, Michael and Matsa, Katerina Eva and Forman-Katz, Naomi and Naseer, Sarah and {St. Aubin}, Christopher and Shearer, Elisa},
  institution = {{Pew Research Center}},
  year = {2024},
  address = {{Washington, D.C.}},
  url = {https://www.pewresearch.org/wp-content/uploads/sites/20/2024/08/PJ_2024.08.29_local-crime-news_report.pdf},
  language = {en-US}
}

@techreport{gallup2023more,
    author  = {Jones, Jeffrey M},
    date    = {2023-11-16},
    title   = {More {A}mericans See {U.S.} Crime Problem as Serious},
    institution = {Gallup},
    address = {{Washington, D.C.}},
    url     = {https://news.gallup.com/poll/544442/americans-crime-problem-serious.aspx}
}

@techreport{gallup2022record,
    author  = {Brenan, Megan},
    date    = {2022-10-28},
    title   = {Record-High {56\%} in {U.S.} Perceive Local Crime Has Increased},
    institution = {Gallup},
    address = {{Washington, D.C.}},
    url     = {https://news.gallup.com/poll/404048/record-high-perceive-local-crime-increased.aspx}
}

@article{nadai2020socio,
    title={Socio-economic, built environment, and mobility conditions associated with crime: a study of multiple cities},
    author={{De Nadai}, Marco and Xu, Yanyan and Letouz{\'e}, Emmanuel and Gonz{\'a}lez, Marta C and Lepri, Bruno},
    journal= {{Scientific Reports}},
    volume={10},
    number={1},
    pages={13871},
    year={2020},
    publisher={Nature Publishing Group UK London},
    doi = {10.1038/s41598-020-70808-2}
}

@unpublished{harcourt2015BWtalk,
    author = {Harcourt, Bernard E.},
    title = {Broken Windows, Again?},
    subtitle = {The problem of Race and Policing in {NYC}},
    date = {2015-03-30},
    note = {Summary accessed from Columbia Law School Story Archive},
    url = {https://www.law.columbia.edu/news/archive/shattering-broken-windows}
}

@book{skogan1990,
    title = {Disorder and decline},
    subtitle = {Crime and the spiral of decay in {American} neighborhoods},
    author = {Skogan, Wesley G.},
    year = {1990},
    publisher = {Free Press},
}

@book{harcourt2001illusion,
    title = {Illusion of order: the false promise of broken windows policing},
    author = {Harcourt, Bernard E.},
    year = {2001},
    publisher = {Harvard University Press},
}

@article{welsh2015meta,
    title = {Reimagining Broken Windows: From Theory to Policy},
    author = {Welsh, Brandon C. and Braga, Anthony A. and Bruinsma, Gerben J. N.},
    journal = {{Journal of Research in Crime and Delinquency}},
    volume={52},
    number={4},
    pages={447--463},
    year={2015},
    doi = {10.1177/0022427815581399}
}

@article{wheeler2021,
    title = {Mapping the Risk Terrain for Crime Using Machine Learning},
    author = {Wheeler, Andrew P. and Steenbeek, Wouter},
    journal = {{Journal of Quantitative Criminology}},
    volume={37},
    number = {},
    pages={445--480},
    year={2021},
    doi = {10.1007/s10940-020-09457-7}
}

@article{zhang2020,
    title={Comparison of Machine Learning Algorithms for Predicting Crime Hotspots}, 
    author={Zhang, Xu and Liu, Lin and Xiao, Luzi and Ji, Jiakai},
    journal= {{IEEE Access}}, 
    volume={8},
    number={},
    pages={181302--181310},
    year={2020},
    doi = {10.1109/ACCESS.2020.3028420}
}

@inproceedings{kim2018,
    author={Kim, Suhong and Joshi, Param and Kalsi, Parminder Singh and Taheri, Pooya},
    booktitle={2018 IEEE 9th Annual Information Technology, Electronics and Mobile Communication Conference (IEMCON)}, 
    title={Crime Analysis Through Machine Learning}, 
    year={2018},
    volume={},
    number={},
    pages={415--420},
}

@article{south2018effect,
    title={Effect of greening vacant land on mental health of community-dwelling adults: a cluster randomized trial},
    author={South, Eugenia C and Hohl, Bernadette C and Kondo, Michelle C and MacDonald, John M and Branas, Charles C},
    journal= {{JAMA Network Open}},
    volume={1},
    number={3},
    pages={e180298--e180298},
    year={2018},
    publisher={American Medical Association},
    doi = {10.1001/jamanetworkopen.2018.0298}
}

@article{cui2022matching,
    author = {Cui, Jesse and Jensen, Shane T and MacDonald, John M},
    title ={The effects of vacant lot greening and the impact of land use and business presence on crime},
    journal = {{Environment and Planning B: Urban Analytics and City Science}},
    volume = {49},
    number = {3},
    pages = {1147-1158},
    year = {2022},
    doi = {10.1177/23998083211050647},
    URL = {https://doi.org/10.1177/23998083211050647},
    eprint = { https://doi.org/10.1177/23998083211050647}
}

@article{corman2005carrots,
    title={Carrots, Sticks, and Broken Windows},
    author={Corman, Hope and Mocan, Naci},
    journal= {{Journal of Law and Economics}},
    volume={48},
    number={1},
    pages={235--266},
    year={2005},
    doi = {10.3386/w9061}
}

@article{harcourt2006experiment,
    title={Broken Windows: New Evidence from {New York City} and a Five-City Social Experiment},
    author={Harcourt, Bernard E. and Ludwig, Jens},
    journal= {{University of Chicago Law Review}},
    volume={73},
    number={1},
    pages={271--320},
    year={2006},
    url = {https://chicagounbound.uchicago.edu/uclrev/vol73/iss1/14/}
}

@techreport{manhattan2001police,
    title = {Do Police Matter? {An} Analysis of the Impact of {New York City’s} Police Reforms},
    author = {Kelling, George L. and {Sousa, Jr.}, William H.},
    institution = {Manhattan Institute for Policy Research},
    year = {2001},
    address = {{New York, NY}},
    url = {https://media4.manhattan-institute.org/pdf/cr_22.pdf},
    language = {en-US}
}

@book{newman1972defense,
    title = {Defensible space},
    subtitle = {Crime prevention through urban design},
    author = {Newman, Oscar},
    year = {1972},
    publisher = {New York, Macmillan},
}

@book{jacobs1961death,
    title = {The Death and Life of Great {American} Cities},
    author = {Jacobs, Jane},
    year = {1961},
    publisher = {New York, Random House},
}

@article{braga2008spots,
    title={Policing crime and disorder hot spots: A randomized controlled trial},
    author={Braga, Anthony A. and Bond, Brenda J.},
    journal= {{Criminology}},
    volume={46},
    number={3},
    pages={577--607},
    year={2008},
    doi = {10.1111/j.1745-9125.2008.00124.x}
}

@article{bernasco2008robberies,
    title={Robberies in {Chicago}: A Block-Level Analysis of the Influence of Crime Generators, Crime Attractors, and Offender Anchor Points},
    author={Bernasco, Wim and Block, Richard},
    journal= {{Journal of Research in Crime and Delinquency}},
    volume={48},
    number={1},
    pages={33--57},
    year={2008},
    doi = {10.1177/0022427810384135}
}

@article{barnum2017kaleidoscope,
    title={The crime kaleidoscope: A cross-jurisdictional analysis of place features and crime in three urban environments},
    author={Barnum, Jeremy D. and Caplan, Joel M. and Kennedy, Leslie W. and  Piza, Eric L.},
    journal= {{Applied Geography}},
    volume={79},
    number={},
    pages={203--211},
    year={2017},
    doi = {10.1016/j.apgeog.2016.12.011}
}

@article{skogan1977changing,
    title={The Changing Distribution of Big-City Crime: A Multi-City Time-Series Analysis},
    author={Skogan, Wesley G.},
    journal= {{Urban Affairs Quarterly}},
    volume={13},
    number={1},
    pages={33–-48},
    year={1977},
    doi = {10.1177/107808747701300102}
}

@article{macdonald2018schools,
    title={Do Schools Cause Crime in Neighborhoods?},
    subtitle = {Evidence from the Opening of Schools in {Philadelphia}},
    author={MacDonald, John M and Nicosia, Nancy and Ukert, Benjamin David},
    journal= {{Journal of Quantitative Criminology}},
    volume={34},
    number={},
    pages={717–-740},
    year={2018},
    doi = {10.1007/s10940-017-9352-y}
}

@article{stec2018forecasting,
  title={Forecasting crime with deep learning},
  author={Stec, Alexander and Klabjan, Diego},
  journal= {{arXiv preprint arXiv:1806.01486}},
  year={2018}
}

@article{kang2017prediction,
  title={Prediction of crime occurrence from multi-modal data using deep learning},
  author= {{Kang, Hyeon-Woo and Kang, Hang-Bong}},
  journal={PloS One},
  volume={12},
  number={4},
  pages={e0176244},
  year={2017},
  publisher={Public Library of Science San Francisco, CA USA}
}

@article{flaxman2019scalable,
author = {Seth Flaxman and Michael Chirico and Pau Pereira and Charles Loeffler},
title = {{Scalable high-resolution forecasting of sparse spatiotemporal events with kernel methods: A winning solution to the NIJ “Real-Time Crime Forecasting Challenge”}},
volume = {13},
journal = {{The Annals of Applied Statistics}},
number = {4},
publisher = {Institute of Mathematical Statistics},
pages = {2564 -- 2585},
year = {2019},
doi = {10.1214/19-AOAS1284},
URL = {https://doi.org/10.1214/19-AOAS1284}
}

@article{rotaru2022event,
  title={Event-level prediction of urban crime reveals a signature of enforcement bias in US cities},
  author={Rotaru, Victor and Huang, Yi and Li, Timmy and Evans, James and Chattopadhyay, Ishanu},
  journal= {{Nature Human Behaviour}},
  volume={6},
  number={8},
  pages={1056--1068},
  year={2022},
  publisher={Nature Publishing Group UK London}
}

@article{kondo2016effects,
  title={Effects of greening and community reuse of vacant lots on crime},
  author={Kondo, Michelle C and Hohl, Bernadette and Han, SeungHoon and Branas, Charles},
  journal= {{Urban Studies}},
  volume={53},
  number={15},
  pages={3279--3295},
  year={2016},
  publisher={Sage Publications Sage UK: London, England}
}

@article{kondo2018blight,
  title={Blight abatement of vacant land and crime in New Orleans},
  author={Kondo, Michelle C and Morrison, Christopher and Jacoby, Sara F and Elliott, Liana and Poche, Albert and Theall, Katherine P and Branas, Charles C},
  journal= {{Public Health Reports}},
  volume={133},
  number={6},
  pages={650--657},
  year={2018},
  publisher={SAGE Publications Sage CA: Los Angeles, CA}
}

@article{heinze2018busy,
  title={Busy streets theory: The effects of community-engaged greening on violence},
  author={Heinze, Justin E and Krusky-Morey, Allison and Vagi, Kevin J and Reischl, Thomas M and Franzen, Susan and Pruett, Natalie K and Cunningham, Rebecca M and Zimmerman, Marc A},
  journal= {{American Journal of Community Psychology}},
  volume={62},
  number={1-2},
  pages={101--109},
  year={2018},
  publisher={Wiley Online Library}
}

@article{branas2018citywide,
  title={Citywide cluster randomized trial to restore blighted vacant land and its effects on violence, crime, and fear},
  author={Branas, Charles C and South, Eugenia and Kondo, Michelle C and Hohl, Bernadette C and Bourgois, Philippe and Wiebe, Douglas J and MacDonald, John M},
  journal= {{Proceedings of the National Academy of Sciences}},
  volume={115},
  number={12},
  pages={2946--2951},
  year={2018},
  publisher={National Acad Sciences}
}

@misc{ACS5year,
    author = {{U.S. Census Bureau}},
    title = {{American Community Survey} 5-year data},
    year = {2024}, 
    url = {data.census.gov},
    urldate = {2024-06-04}
}

@misc{censusGeo,
    author = {{U.S. Census Bureau}},
    title = {{TIGER/Line} Shapefiles},
    year = {2024}, 
    url = {https://www.census.gov/geographies/mapping-files.html},
    urldate = {2024-06-04}
}

@misc{chicago2024crime,
    author = {{Chicago Police Department}},
    title = {Crimes - 2001 to Present},
    year={2024},
    url = {data.cityofchicago.org},
    urldate = {2024-06-04}
}

@misc{nyc2024crime,
    author = {{New York Police Department}},
    title = {{NYPD} Complaint Data Current (Year To Date)},
    year={2024},
    url = {data.cityofnewyork.us},
    urldate = {2024-06-04}
}

@misc{healthychicago,
    author = {{Chicago Department of Public Health}},
    title = {{Healthy Chicago Survey}},
    year = {2025}, 
    url = {https://www.chicago.gov/city/en/depts/cdph/supp_info/healthy-communities/healthy-chicago-survey.html},
    urldate = {2024-06-04}
}

\end{document}